\documentclass[journal]{IEEEtran}

\ifCLASSINFOpdf
\else
\fi
\usepackage{balance}
\usepackage{amsmath,epsfig}
\usepackage{lineno}
\usepackage{enumerate}
\usepackage{amsfonts}
\usepackage{times}
\usepackage{helvet}
\usepackage{courier}
\usepackage{multirow}
\usepackage{amssymb}
\usepackage{graphicx}
\usepackage{subfigure}
\usepackage{longtable}
\usepackage{rotating}
\usepackage{multicol}
\usepackage{balance}
\usepackage{array}
\usepackage{tabularx}
\usepackage{tabu}
\usepackage{makecell}
\usepackage{cite}

\begin{document}
%
\title{Locality Preserving Joint Transfer \\for Domain Adaptation}
%
%
%

\author{Jingjing Li,
        Mengmeng Jing,
        Ke Lu,
        Lei Zhu,
        Heng Tao Shen
\thanks{J. Li, M. Jing, K. Lu and H. T. Shen are with the School
of Computer Science and Engineering, University of Electronic Science and Technology of China;
Lei Zhu is with Shandong Normal University.
E-mail: jjl@uestc.edu.cn. }}

\markboth{IEEE Transactions on Image Processing}%
{Shell \MakeLowercase{\textit{et al.}}: Bare Demo of IEEEtran.cls for Journals}

\maketitle

\begin{abstract}
Domain adaptation aims to leverage knowledge from a well-labeled source domain to a poorly-labeled target domain. A majority of existing works transfer the knowledge at either feature level or sample level. Recent researches reveal that both of the paradigms are essentially important, and optimizing one of them can reinforce the other. Inspired by this, we propose a novel approach to jointly exploit feature adaptation with distribution matching and sample adaptation with landmark selection. During the knowledge transfer, we also take the local consistency between samples into consideration, so that the manifold structures of samples can be preserved. At last, we deploy label propagation to predict the categories of new instances. Notably, our approach is suitable for both homogeneous and heterogeneous domain adaptation by learning domain-specific projections. Extensive experiments on five open benchmarks, which consist of both standard and large-scale datasets, verify that our approach can significantly outperform not only conventional approaches but also end-to-end deep models. The experiments also demonstrate that we can leverage handcrafted features to promote the accuracy on deep features by heterogeneous adaptation. {\it Our codes are available at https://github.com/lijin118/LPJT}.
\end{abstract}


\IEEEpeerreviewmaketitle

\section{Introduction}
\IEEEPARstart{T}{he} scarcity principle\footnote{The scarcity principle is an economic theory in which a limited supply of a good, coupled with a high demand for that good, results in a mismatch between the desired supply and demand equilibrium. (Investopedia)} is one of the commonsense principles in our daily life. In the field of machine learning, the scarcity shows up in labeled samples. In most cases, labeling samples is labor intensive. Or, in other cases, it is even hard to collect samples. However, most of the effective machine learning approaches need to be trained in a supervised fashion. To address this mismatch, transfer learning~\cite{pan2010survey,long2014graph,ding2015deep} has been proposed to borrow knowledge from related and yet well-labeled domains. Unfortunately, one cannot directly transfer from one domain (the source domain) to another (the target domain) due to the distribution divergence~\cite{pan2010survey,ding2017robust}. We have to align the source domain and the target domain to secure the knowledge transfer since a basic assumption of statistical learning is that the training set and test set are drawn from the same data distribution~\cite{pan2010survey}. Recently, a battery of researches have been proposed to align the source domain and the target domain in terms of domain adaptation~\cite{bruzzone2010domain,gopalan2011domain,ganin2014unsupervised,lu2018embarrassingly,zhang2016robust,saito2018maximum,pinheiro2018unsupervised}.

The basic assumption behind domain adaptation is that some latent factors are shared by the source domain and the target domain. These latent factors can be specific features~\cite{li2014learning}, geometric structures~\cite{zhang2017joint} and landmark samples~\cite{aljundi2015landmarks}. For clarity, we give a brief introduction for the three aspects. Firstly, since the source domain and the target domain share the same semantic space in domain adaptation, sample features can be decomposed into shared and domain-dependent features. The specific features in this paper refer to the shared features which are generally uncovered by a shared subspace. In addition, high dimensional data lies on a low dimensional manifold embedded in the ambient space~\cite{he2005face,cai2007spectral,yan2007graph}. The geometric structures refer to the manifold structures. At last, the landmark samples are a subset of labeled data instances in the source domain that are distributed most closely to the target domain~\cite{gong2013connecting}. The very essential idea of existing domain adaptation methods is to reveal these latent factors and use them as the bridge of knowledge transfer. 

\begin{figure*}[t]
\begin{center}
\subfigure[Samples in the original space]{
 \includegraphics[width=0.21\linewidth]{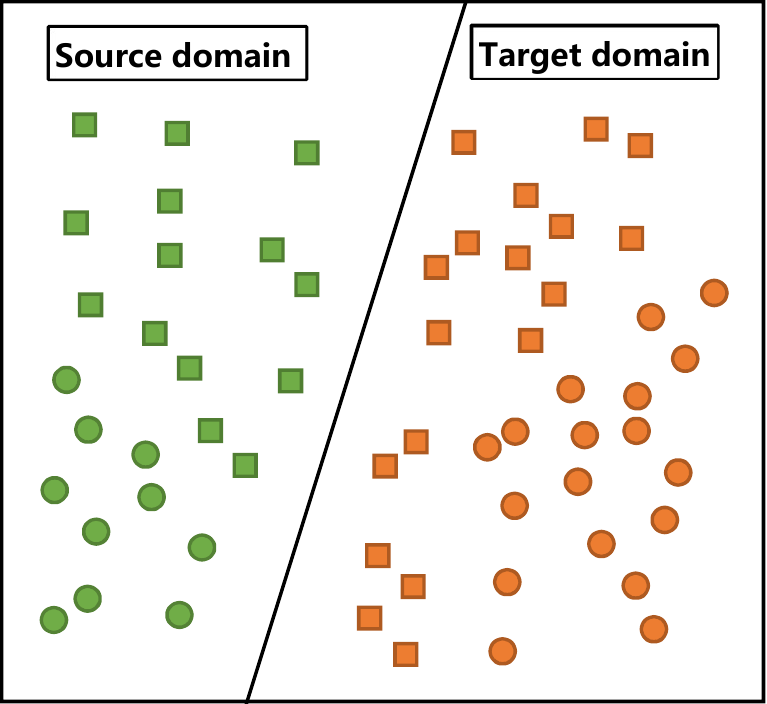}
}
\subfigure[Distribution matched samples]{
  \includegraphics[width=0.21\linewidth]{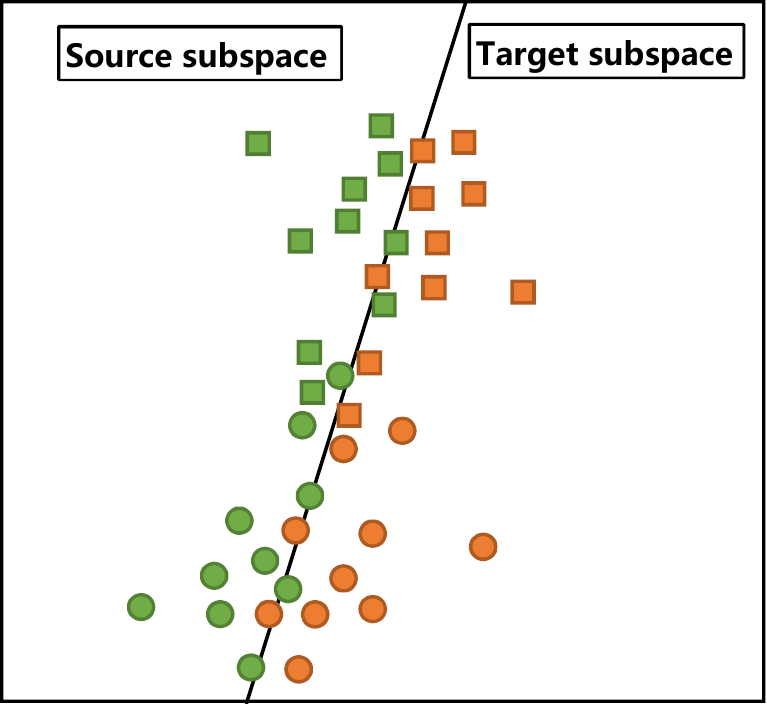}
}
\subfigure[Sample re-weighting]{
 \includegraphics[width=0.21\linewidth]{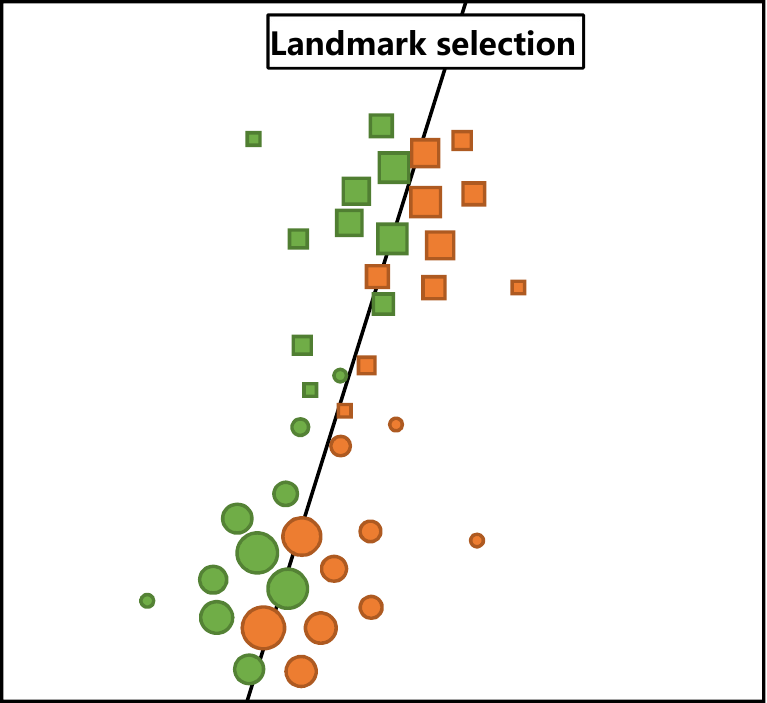}
}
\vspace{-5pt}
\caption{The main idea of our approach. Colors denote domains and shapes represent categories. The size of a point indicates the weight of a sample. (a) Distribution divergence of data in their original spaces is large; (b) Distribution divergence between domains is minimized after mapping data into a latent shared subspace; (c) Selecting landmarks that can bridge the source and target subspaces. }
\label{fig:main}
\end{center}
\vspace{-15pt}
\end{figure*}

Recently, several researches~\cite{li2018transfer,hubert2016learning,zhang2017joint,li2016joint} reveal that jointly optimizing more than one of the three latent factors can be an effective way to handle domain adaptation. For instance, Aljundi et al.~\cite{aljundi2015landmarks} introduce an approach based on both subspace alignment and landmark selection. Hubert et al.~\cite{hubert2016learning} propose to select landmarks and match feature distributions at the same time. Long et al.~\cite{long2014transfer} explore the benefits from jointly optimizing feature alignment and sample~re-weighting. 

Previous approaches have proved that each of the adaptation paradigms is essentially important, and optimizing one of them can reinforce the other. However, there are still deficiencies in existing work. The most conspicuous two deficiencies are: 1) The joint optimization is insufficient. Specifically, although existing work have investigated joint optimization, a vast majority of them~\cite{long2014transfer,ding2017robust,zhang2017joint} only jointly optimize two of the three and let the other one be neglected. 2) The existing models are limited in application scenarios. Specifically, most of existing work~\cite{gong2012geodesic,gopalan2011domain,pan2011domain,zhang2017joint} are limited to only homogeneous domain adaptation, they cannot be directly applied on heterogeneous domain adaptation problems~\cite{hubert2016learning}. The first deficiency can be attributed to the challenge of simultaneously handling three concerns in one objective. The second deficiency is quite common in real-world applications since we cannot always guarantee that the source domain and the target domain are sampled with homogeneous features. 

In this paper, we propose a novel approach to overcome the two shortcomings in a unified objective. On one hand, to take full advantage of joint optimization, we align features with distribution matching and re-weight samples with landmark selection. Our idea is shown in Fig.~\ref{fig:main}. In addition, we also take the sample geometric structure into consideration, so that the local manifold structures of samples can be preserved during the knowledge transfer. Different from existing approaches which only investigate the local structure in feature learning, we take a step further and introduce label propagation~\cite{zhu2002learning} to maximize the utilization of locality preservation. On the other hand, to ensure that our model can be applied to domains with arbitrary feature dimensionalities, we propose to learn two different mappings, one for each domain, to project the two domains into a latent space. With an ingenious formulation, the two mappings are inter-connected and learned together, so that the latent space will be shared by the two domains and in which the two domains can be well aligned. In summary, the main contributions of this work can be outlined as follows:

\begin{enumerate}[1)]
\item We propose a novel domain adaptation method that considers knowledge transfer at both feature level and sample level. Specifically, at the feature level, our method can reduce the distribution divergence between domains. At the sample level, our method can preserve the neighborhood relationship of samples and make it robust to outliers. The proposed method takes full advantage of joint optimization, overcomes the challenges of formulation and optimization, and shows significant advantage against previous approaches, even end-to-end deep models. 

\item Different from previous approaches which are limited to specific application scenarios, our model is more general. By simultaneously optimizing two projections, one for each domain, our method can handle domain adaptation problems where the source and the target domains have different dimensionalities and features. It suits for both homogeneous and heterogeneous domain adaptation, either in unsupervised or supervised learning manner. 

\item In the experiments, we do not only put our eyes on verifying the effectiveness of the proposed method. We are interested to see if the handcrafted features can assist the task on deep features, and if different deep features can facilitate each other. We verify that features extracted from one network can reinforce the others, and even handcrafted features can boost the accuracy on deep features by heterogeneous domain adaptation, which highlights the value of transfer learning. 
\end{enumerate}


\section{Related Work}

\subsection{Domain Adaptation}
Domain adaptation~\cite{gong2012geodesic,long2014transfer,pan2011domain,li2016joint,li2017structured} is a popular research branch of transfer learning~\cite{pan2010survey,zhu2018exploring}, which aims to leverage knowledge from a well-labeled domain to a poor-labeled but related one. In general, the well-labeled source of knowledge transfer is called source domain, and the poor-labeled target of knowledge transfer is called target domain. According to the label conditions, domain adaptation can be grouped as unsupervised domain adaptation~\cite{fernando2013unsupervised,long2014transfer} and semi-supervised domain adaptation~\cite{li2014learning,hubert2016learning}. Unsupervised domain adaptation handles the problem where no labeled target samples are available. Semi-supervised domain adaptation is applicable for tasks where a few labeled target samples are involved in the training stage. Our proposed method can work in both unsupervised and semi-supervised manners.

According to the property of sample features, domain adaptation can be categorized as homogeneous domain adaptation~\cite{gopalan2011domain,gong2013connecting,jing2018learning} and heterogeneous domain adaptation~\cite{hubert2016learning,li2018transfer,li2018heterogeneous}. In homogeneous domain adaptation, the data from both domains are sampled from the same feature space. As an extension of homogeneous domain adaptation, heterogeneous methods can handle domains with arbitrary features and dimensionalities. Specifically, multi-source domain adaptation~\cite{gong2013reshaping,hoffman2012discovering,liu2018coupled,ding2016transfer} handles the problems where more than two domains are involved in the task.

According to the learning mechanisms, existing domain adaptation approaches can be roughly classified as feature-based methods~\cite{pan2011domain,long2013transfer,li2018transfer}, sample-based methods~\cite{aljundi2015landmarks,gong2013connecting}, classifier-based methods~\cite{li2014learning,li2018domain} and neural network-based methods~\cite{tzeng2017adversarial,hoffman2017cycada}.

Feature-based methods can be further categorized as distribution matching, e.g., subspace learning~\cite{pan2011domain,long2013transfer}, and property preserving, e.g., geometrical preserving~\cite{zhang2017joint}. Distribution matching algorithms reduce marginal or conditional distribution divergence through learning a latent subspace shared by the two domains. The MMD-based subspace learning algorithms~\cite{pan2011domain,long2013transfer} are the typical examples of distribution matching. For instance, TCA~\cite{pan2011domain} learns some transferable components across domains in a Reproducing Kernel Hilbert Space using MMD which mitigates the marginal distribution between domains. JDA~\cite{long2013transfer} jointly considers the marginal and the conditional distributions and builds a new feature representation to guarantee robustness when the distribution divergence is large. Property preserving algorithms preserve important properties, e.g., geometric structure~\cite{gong2012geodesic}, when the samples are mapped into a low-dimensional subspace. Sample-based methods generally handle the domain adaptation by landmark selection~\cite{gong2013connecting}. CDLS~\cite{hubert2016learning}, for instance, selects landmarks that have higher matching degrees across domains by minimizing the MMD distance. LSSA~\cite{aljundi2015landmarks} uses the Gaussian kernel to compute quality measures of all samples and selects landmarks whose quality measures are above a threshold. 

Nowadays, most of the state-of-the-art results are achieved by deep neural network-based models~\cite{ganin2014unsupervised,long2015learning,rozantsev2018beyond}. These deep methods can be taxonomically divided into three categories: 1) discrepancy-based methods~\cite{long2015learning,tzeng2014deep,ghifary2016deep,zhuang2015supervised} which minimize the distribution discrepancy between the source domain and the target domain; 2) adversarial-based methods~\cite{tzeng2017adversarial,hoffman2017cycada,zhang2018collaborative,saito2018maximum} which optimize a two-player minimax game; 3) reconstruction-based methods~\cite{bousmalis2016domain,ghifary2016deep,yi2017dualgan,zhu2017unpaired,kim2017learning} which discover and utilize the cross-domain relations. Compared with conventional domain adaptation approaches, end-to-end deep models integrate feature learning and knowledge adaptation into a consistent operation. Generally, the deep models also show better performance than conventional ones. 

\subsection{Preliminaries}
{\bf Graph Embedding}. Graph embedding~\cite{yan2007graph} is a general framework for dimensionality reduction. The main idea of graph embedding is to preserve the sample relationships during dimensionality reduction. For instance, it minimizes the intra-class distances and maximizes the inter-class distances so that the learned features are discriminative. In graph embedding, samples are formulated as a graph. The data points are nodes and the distances among them are edge weights. If we deploy Gaussian distance to calculate the weight $W$, we have:
\begin{equation}  
W_{ij}\!\!=\!\!\begin{cases}
                      {\rm exp}(\frac{- (||x_i - x_j||^2)}{2}), & \!\!\!\!\mbox{if nodes $x_i$ and $x_j$ are connected} \\ 
                      0 &  \!\!\!\!\mbox{otherwise}. \end{cases}
\end{equation}   
Then, a graph Laplacian $L$ can be calculated by $L=D-W$, where $D$ is a diagonal matrix with ${D}_{ii}=\sum_{j\neq i}{W}_{ij}$. At last, the dimensionality reduction can be formulated as optimizing two graph Laplacians. More details~can~be~found~in~\cite{yan2007graph}.

{\bf Label Propagation}. Label propagation~\cite{zhu2002learning} investigates the problem of using unlabeled data to facilitate the classification of labeled data. The main assumption behind label propagation is that close data points tend to have close labels. In label propagation, all data points are formulated as nodes of a fully connected graph, and the edge weight between two nodes is calculated by the Euclidean distance of them. Label propagation works in an iterative manner. If we use $Y(t)$ to indicate the labels corresponding to time $t$, the labels in time $t+1$ can be optimized by~\cite{zhou2004learning}:
\begin{equation}  
Y(t+1)= \sigma SY(t) + (1-\sigma)Y(0),
\end{equation} 
where $\sigma \in (0,1)$ is a parameter, $S=D^{-1/2}WD^{-1/2}$ is a similarity matrix. The definitions of $W$ and $D$ are same with that introduced in graph embedding. More details of the algorithm can be found in~\cite{zhou2004learning}.  

\section{Locality Preserving Joint Transfer}

\subsection{Definitions and Notations}

    {\bf Definition 1.}~A domain $D$ is defined by a feature space $\chi$ and its probability distribution $P(X)$, where $X \in \chi$. For a specific domain, a classification task $T$ consists of class information $y$ and a classifier $f(x)$, that is $T = \{y, f(x)\}$.
    We use subscripts $s$ and $t$ to indicate the source domain and the target domain, respectively. This paper focuses on the following problem:

    {\bf Problem 1.} Given a labeled source domain $\{X_{\rm s}, Y_{\rm s}\}$ and an unlabeled target domain $X_{\rm t}$, where $X_{\rm s}$ and $X_{\rm t}$ are the source and the target domain samples, $Y_{\rm s}$ is the labels of the source domain, $P(X_{\rm s}) \neq P(X_{\rm t})$ and $P(y_{\rm s}|X_{\rm s}) \neq P(y_{\rm t}|X_{\rm t})$, we aim to project the source and the target domain into a subspace by projection matrices $A$ and $B$ so that the common latent features shared by the involved domains are uncovered, the data manifold structures are preserved, and the domain gaps are minimized.

 For clarity, we show the frequently used notations and corresponding descriptions in Table~\ref{tab:notaions}.

\begin{table}[t]
\centering
\caption{Notations and corresponding descriptions.}
\vspace{-5pt}
\small
\begin{tabular}{c|l}
\Xhline{0.75pt}
Notations & Descriptions \\ \hline
    ${E}_{MG}$      &       The marginal cross-domain data distributions       \\ 
     ${E}_{CD}$     &       The conditional cross-domain data distributions       \\ 
      ${X}_s$    &   The source domain data           \\ 
   ${X}_u$       &   The unlabeled target domain data             \\ 
   ${A}$       &    The source domain projection matrix          \\ 
    ${B}$      &    The target domain projection matrix          \\ 
    ${n}_s$      &  The number of source domain data            \\ 
    ${n}_u$      &   The number of unlabeled target domain samples          \\ 
     $\alpha$     &     The weight vector for source domain data         \\ 
      $\beta$          &   The weight vector for target domain data           \\ 
     $\delta$      &  The average weight of both domains                \\ 
      ${H}_{sm}$   &      Coefficient matrix                   \\ 
 $L_{w}$ &  The Laplacian matrix of the intrinsic graph  \\ 
   $L_{b}$  & The Laplacian matrix of the penalty graph  \\ 
       ${S}^u_h$  & The covariance matrix of the target domain  \\ 
      $\gamma$  &Trade-off parameter for locality preserving  \\ 
       $\mu$  &Trade-off parameter for target variance \\ \Xhline{0.75pt}                                                      
\end{tabular}
\label{tab:notaions}
\vspace{-10pt}
\end{table}

\subsection{Problem Formulation}
\label{sec:formulation}
In this paper, we handle the domain adaptation problem by learning a subspace where the common latent features shared by involved domains are uncovered, the data manifold structures are preserved, and the domain shifts are minimized. To this end, we formulate our objective as:
\begin{equation}  \underset{~}{\mbox{min}}\;{\frac 
        {{\it \{domain~shift\}}_{\rm MG}+{\it \{domain~shift\}}_{\rm CD}+{\it \{dist\}}_{intra}}
        {{\it \{dist\}}_{inter}}} .   \label{eq:obj_abstract} \end{equation} 

In Eq.~\eqref{eq:obj_abstract}, ${\it \{domain~shift\}}_{\rm MG}$ and ${\it \{domain~shift\}}_{\rm CD}$ indicate the marginal and conditional distributions, respectively, which are used to reduce the domain shifts; ${\it \{dist\}}_{intra}$ and ${\it \{dist\}}_{inter}$ represent the intra-class sample distance and inter-class sample distance, respectively, which are used to preserve the local manifold structures. Our method handles domain adaptation at both feature level and sample level. At the feature level, two mapping functions $A$ and $B$ are learned to project the two domains into a shard space so that the common latent features can be uncovered. At the sample level, two weight vectors $\alpha$ and $\beta$ are learned to select landmarks. The details of each part are presented in the following subsections. \\

    \subsubsection{Distribution Matching}
      Our approach reveals the shared latent factors by learning a domain-invariant subspace where the two domains can be well aligned at both feature level and sample level. Previous works~\cite{long2013transfer,gu2011joint} only learn one domain-shared projection, which suffers a vital limitation when the source domain and the target domain have different dimensionalities. To gain better generalization ability, we learn two different projection matrices, one for each domain. We deploy these two matrices to project the source and the target domain data into a latent shared subspace. Specifically, let $A$ be the projection matrix for the source domain and $B$ for the target domain. $X_{\rm s}\, \in {\mathbb R}^{d_s \times n_{\rm s}}$ and $X_{\rm t}\, \in {\mathbb R}^{d_t \times n_{\rm t}}$ are the source and the target domain data respectively, where $n_{\rm s}$ and $n_{\rm t}$ are the total numbers of the corresponding domain samples, $d_s$ and $d_t$ are their dimensionalities. $A \in {\mathbb R}^{d_s \times d}$ can project the source domain data into a $d$-dimensional subspace, where $d \ll \min(d_s,d_t)$. Then, the low-dimensional data can be represented by $A^{\rm T} X_{\rm s}$. Similarly, the low-dimensional representation of the target domain data $X_{\rm t}$ can be represented as $B^{\rm T} X_{\rm t}$, and the low-dimensional representation of the {\it unlabeled} target domain data $X_{\rm u}$ can be $B^{\rm T} X_{\rm u}$. Considering the large distribution divergence, we align both the marginal and the conditional distributions between the two domains by adopting MMD~\cite{long2014transfer,gretton2006kernel}, which can be formulated as:
      \begin{equation} \underset{A,B}{\mbox{min}}\; E_{\rm MG}(X_{\rm s}, X_{\rm u}, A, B) + E_{\rm CD}(X_{\rm s}, X_{\rm u}, A, B) \enspace , \label{eq:dist_matct_ini} \end{equation} 
      \noindent where $X_{\rm s}$ is the source samples, $X_{\rm u}$ is the {\it unlabeled} target samples. It is worth noting that several existing approaches~\cite{hubert2016learning,li2014learning} split $X_t$ into unlabeled $X_u$ and labeled $X_l$ and use both of them for training. We only use $X_u$ to learn the model. $E_{\rm MG}$ and $E_{\rm CD}$ match the marginal and the conditional cross-domain data distributions, respectively. For simplicity, we set $E_{\rm MG}$ and $E_{\rm CD}$ with the same weight. In detail, the marginal MMD $E_{\rm MG}$ can be calculated by:     
     \begin{equation} \begin{array}{c}E_{\rm MG} = \Big|\!\Big|{\frac{1}{n_{\rm s}}}\sum^{n_{\rm s}}_{i=1}A^{\rm T}x^i_{\rm s} - {\frac{1}{n_{\rm u}}\sum^{n_{\rm u}}_{j=1} B^{\rm T}x^j_{\rm u}\Big|\!\Big|^2} {\enspace ,}  \label{eq:em_no_landmark}  \end{array} \end{equation} 
      and the conditional MMD $E_{\rm CD}$ can be calculated by:
      \begin{equation} 
      \label{eq:ec_no_landmark}
      \begin{array}{c}
      E_{\rm CD} = \, \sum\limits^C_{c=1} \Big|\!\Big|{\frac{1}{n^c_{\rm s}}}\sum\limits^{n^c_{\rm s}}_{i=1} A^{\rm T} x^{i,c}_{\rm s} - {\frac{1}{n^c_{\rm u}}}\sum\limits^{n^c_{\rm u}}_{j=1} B^{\rm T} x^{j,c}_{\rm u}\Big|\!\Big|^2 +\\ 
      \,~~~~~~ \sum\limits^C_{c=1} ({\frac{1}{n^c_{\rm s} n^c_{\rm u}}}\sum\limits^{n^c_{\rm s}}_{i=1}\sum\limits^{n^c_{\rm u}}_{j=1}\Big|\!\Big| A^{\rm T} x^{i,c}_{\rm s} - B^{\rm T} x^{j,c}_{\rm u}\Big|\!\Big|^2) ,
      \end{array}
      \end{equation}
      where $C$ is the number of classes, $n^c_{\rm s}$ and $n^c_{\rm u}$ denote the total numbers of the source and the target domain samples in class $c$, respectively. In $E_{\rm CD}$, the first part calculates the discrepancy of class-wise means. The second part preserves the neighborhood relationship which encourages the samples from the same category to be close with each other. Since labels in the target domain are not available, we use the pseudo labels for target domain samples. Specifically, we classify the target domain with label propagation~\cite{zhu2002learning} for initialization in this paper. Then, the labels are iteratively updated during the optimization. For semi-supervised settings, we use the ground-truth label to replace the pseudo label. \\

    \subsubsection{Landmark Selection}
      At the sample level, we learn two weight vectors $\alpha$ and $\beta$ for the source and the target domains, respectively. Each element of the vector is a weight of the corresponding sample. As a result, the objective function can be rewritten as:
      \begin{equation} \begin{array}{c} \underset{A,B}{\mbox{min}}\; E_{\rm MG}(\alpha,\beta,X_{\rm s}, X_{\rm u}, A, B) + E_{\rm CD}(\alpha,\beta,X_{\rm s}, X_{\rm u}, A, B) \enspace , \\  {\mbox{s.t.}}\;\{\alpha^c_i,\beta^c_i\}\in [0,1],{\frac{\alpha^{c^{\rm T}}{{\bf 1}_{n^c_{\rm s} } }}{n^c_{\rm s}}} = {\frac{\beta^{c^{\rm T}}{{\bf 1}_{n^c_{\rm u} } } }{n^c_{\rm u}}} = \delta , \label{eq:landmark_intro} \end{array} \end{equation} 
 where $\alpha = [\alpha^1;\cdots;\alpha^c;\cdots;\alpha^C] \in {\mathbb R}^{n_{\rm s}}$ and $\beta = [\beta^1;\cdots;\beta^c;\cdots;\beta^C] \in {\mathbb R}^{n_{\rm u}}$ are the weights of data in the source domain and the target domain, respectively, $\alpha^c = [\alpha^c_1;\cdots;\alpha^c_{n^c_{\rm s}}]$, $\beta^c = [\beta^c_1;\cdots;\beta^c_{n^c_{\rm u}}]$, ${\bf 1}_{n^c_{\rm s} } \in {\mathbb R}^{n^c_{\rm s}} $ and ${\bf 1}_{n^c_{\rm u} } \in {\mathbb R}^{n^c_{\rm u}} $ are column vectors with all ones. $\delta \in [0,1]$ controls the ratio of landmarks in the whole source or target domain samples. The constraints on $\alpha$ and $\beta$ keep them from trivial solutions such as one-hot vectors that only align one sample from source and one sample from target. Then, $E_{\rm MG}$ and $E_{\rm CD}$ in Eq.(\ref{eq:landmark_intro}) further can be calculated by:      
      \begin{equation}
      \begin{array}{c}
      E_{\rm MG}\, = \,\Big|\!\Big|{\frac{1}{\delta n_{\rm s}}}\sum^{n_{\rm s}}_{i=1}\alpha_i A^{\rm T}x^i_{\rm s} - {\frac{1}{\delta n_{\rm u}}}\sum^{n_{\rm u}}_{j=1}\beta_j B^{\rm T}x^j_{\rm u}\Big|\!\Big|^2 {\enspace ,} \label{eq:em_full_landmark}
      \end{array}
      \end{equation}      
      \begin{equation}
      \begin{array}{c}
       E_{\rm CD}  =  \sum\limits^C_{c=1}  \Big|\!\Big|{\frac{1}{\delta n^c_{\rm s}}}\sum\limits^{n^c_{\rm s}}_{i=1}\alpha_i A^{\rm T} x^{i,c}_{\rm s}\!\! -\!\! {\frac{1}{\delta n^c_{\rm u}}}\sum\limits^{n^c_{\rm u}}_{j=1}\beta_j B^{\rm T} x^{j,c}_{\rm u}\Big|\!\Big|^2 \!\!\\~~~+  
      \sum\limits^C_{c=1} ({\frac{1}{\delta^2 n^c_{\rm s} n^c_{\rm u}}}\sum\limits^{n^c_{\rm s}}_{i=1}\sum\limits^{n^c_{\rm u}}_{j=1}\Big|\!\Big|\alpha_i A^T x^{i,c}_{\rm s}\!\! -\!\! \beta_j B^T x^{j,c}_{\rm u} \Big|\!\Big|^2 ). \!\!\!\!{\enspace }  \label{eq:ec_full_landmark}
       \end{array}
       \end{equation}

      For clarity, We rewrite Eq.~\eqref{eq:em_full_landmark} and Eq.~\eqref{eq:ec_full_landmark} into the following equivalent equations in the form of matrix:
      \begin{equation} \begin{aligned} &E_{\rm MG}=A^T X_s H_{sm} X^T_s A + B^T X_u H_{um} X^T_u B \\ &~~~~~~~~- 2 A^T X_s H_{sum} X^T_u B,  \end{aligned}  \label{eq:matrix_marg_hsm}\end{equation}
      and
      \begin{equation} \begin{aligned} &E_{\rm CD}=A^T X_s H_{sc} X^T_s A + B^T X_u H_{uc} X^T_u B \\ &~~~~~~~~- 2 A^T X_s H_{suc} X^T_u B,   \end{aligned}  \label{eq:matrix_cond_hsc}\end{equation}
      where $H$ with different subscripts are coefficient matrices defined as follows: 
      \[ \begin{aligned}  H_{sm} &= {\frac{1}{\delta^2 n^2_s}} \alpha \cdot \alpha^T,~ H_{um} = {\frac{1}{\delta^2 n^2_u}} \beta \cdot \beta^T,\\ H_{sum} &= {\frac{1}{\delta^2 n_s n_u}} \alpha \cdot \beta^T,\end{aligned}\]

      $H_{sc}, H_{uc}$ and $H_{suc}$ are diagonal matrices with $H^c_{sc} = {\frac{1}{\delta^2 {n^c_s}^2}} \alpha^c {\alpha^c}^T + {\frac{1}{\delta^2 {n^c_s}}}diag(\alpha^c \odot \alpha^c)$, $H^c_{uc} = {\frac{1}{\delta^2 {n^c_u}^2}} \beta^c {\beta^c}^T + {\frac{1}{\delta^2 {n^c_u}}}diag(\beta^c \odot \beta^c)$ and $H^c_{suc} = {\frac{2}{\delta^2 n^c_s n^c_u}} \alpha^c {\beta^c}^T$.

      %

      After some algebra operations, Eq.~\eqref{eq:landmark_intro} can be written as the following equivalent equation:
      \begin{equation} A^T M_{ss} A + B^T M_{uu} B - 2 A^T M_{su} B,  \label{eq:matrix_form_mss} \end{equation}
      with
      \[ \begin{aligned} M_{ss} &= X_s ( H_{sm} + H_{sc} ) X^T_s, \\ M_{uu} &= X_u ( H_{um} + H_{uc} ) X^T_u,\\ M_{su} &= X_s ( H_{sum} + H_{suc} ) X^T_u.   \end{aligned}\]

      At last, the above equation, i.e. (\ref{eq:matrix_form_mss}), can be further transformed to its matrix form as follows:
      \begin{equation} \underset{A,B}{\mbox{min}} \;{\rm Tr}\left(\begin{bmatrix}A^{\rm T} & B^{\rm T}\end{bmatrix}{\begin{bmatrix}M_{\rm ss} & M_{\rm su}\\ M_{\rm us} & M_{\rm uu}\end{bmatrix}}{\begin{bmatrix}A\\B\end{bmatrix}}\right) \enspace ,  \label{eq:trace_matrix_form_mss} \end{equation} 
where ${\rm Tr(\cdot)}$ denotes the trace of a matrix.  Each entry $(H_{\rm ss})_{ij}$ in $H_{\rm ss} \in {\mathbb R}^{n_{\rm s} \times n_{\rm s} } $ denotes the coefficient associated with $ { x^i_{\rm s}}^{\rm T} x^j_{\rm s} $. Similar remarks can be applied to $H_{\rm tt} \in {\mathbb R}^{n_{\rm t} \times n_{\rm t} }$ and $H_{\rm st} \in {\mathbb R}^{n_{\rm s} \times n_{\rm t} }$. \\

\begin{figure}[t]
\begin{center}{}
 \includegraphics[width=0.49\linewidth]{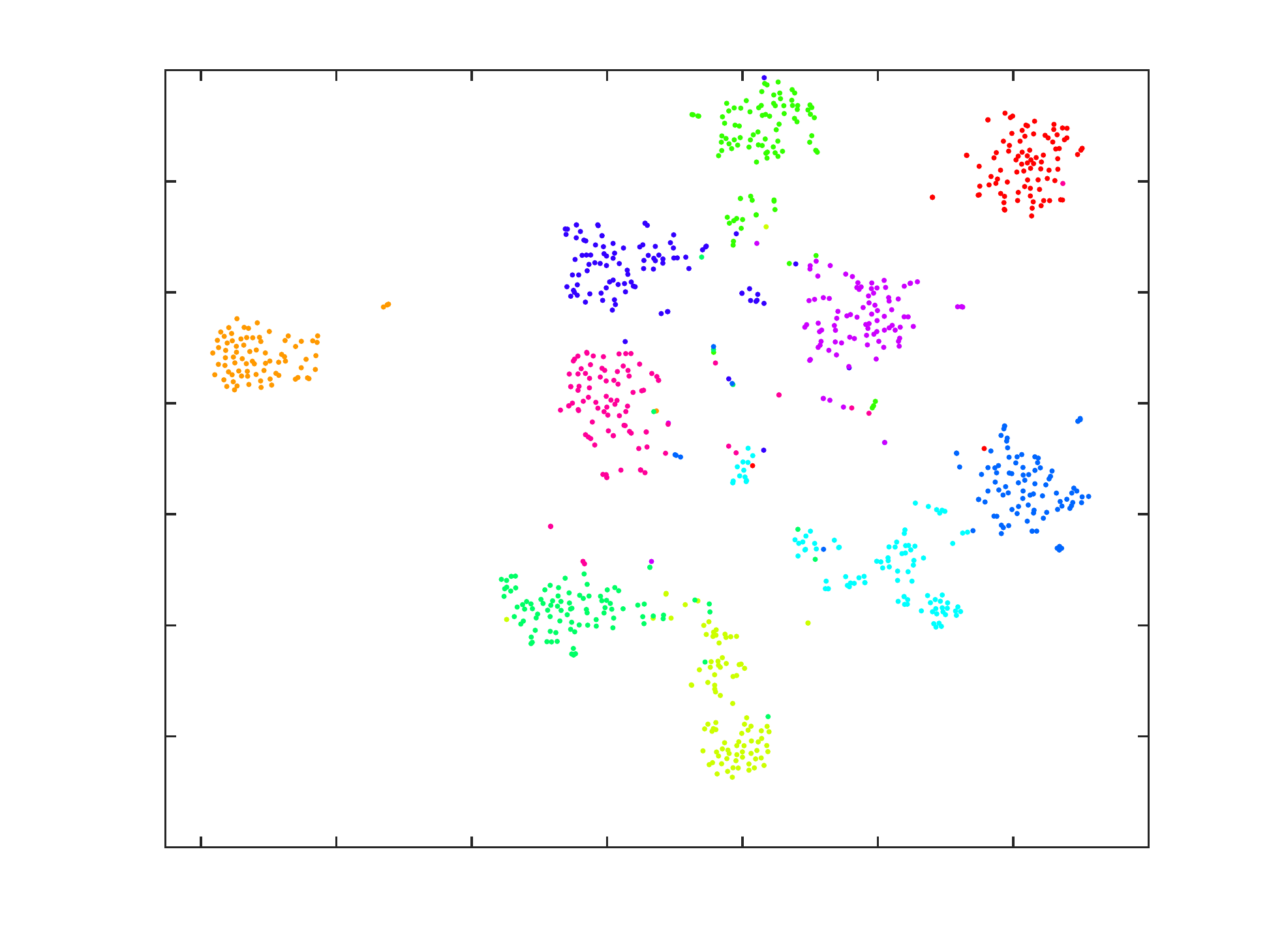}
  \includegraphics[width=0.49\linewidth]{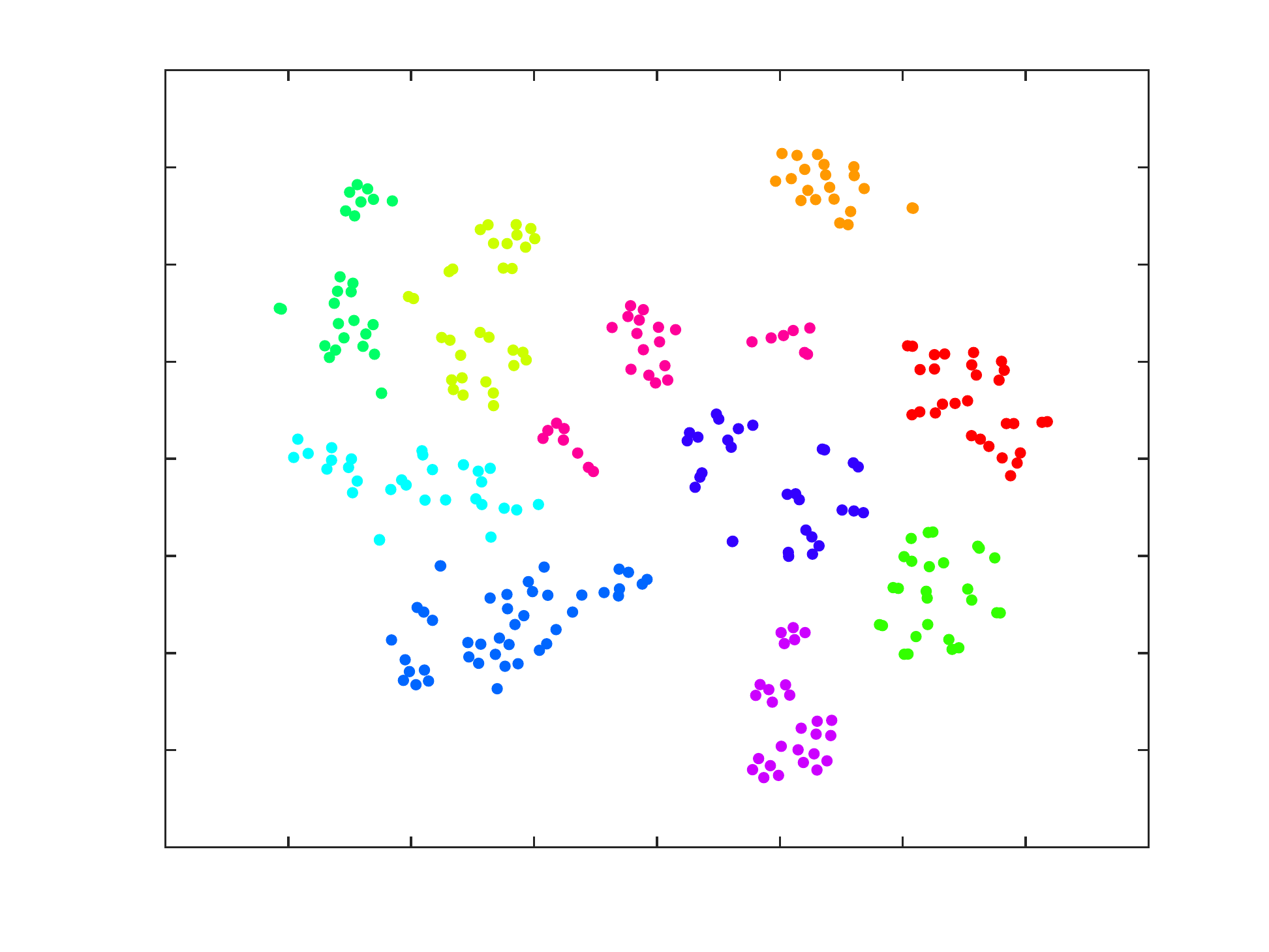}
  \vspace{-20pt}
\caption{Illustration of local consistency. The figures are generated by t-SNE~\cite{maaten2008visualizing} on Amazon and Webcam datasets in which each color represents a category. It is easy to see that samples from the same category tend to be close with each other. In our formulation, we aim to preserve such a neighborhood relationship during the knowledge transfer.}
\label{fig:local}
\end{center}
\vspace{-16pt}
\end{figure}

    \subsubsection{Locality Preservation} The local consistency of sample relationship indicates that a sample tends to have the same label with its $k$-nearest neighbors, as shown in Fig.~\ref{fig:local}. Such a local consistency has been widely discussed in manifold learning~\cite{he2005face,cai2007spectral,yan2007graph}. In transfer learning, it would be helpful if we can encourage samples from the same class stay close during the distribution matching and keep samples far from each other if they are from different classes. To this end, we introduce two Laplacian graph terms, one for each domain, by deploying the Fisher criterion~\cite{yan2007graph}:
      \begin{normalsize} \begin{equation} {\mbox{min}}\,{\frac{{\rm Tr}(A^{\rm T}X_{\rm s}L^{\rm s}_{\rm w}X^{\rm T}_{\rm s}A)}{{\rm Tr}(A^{\rm T}X_{\rm s}L^{\rm s}_{\rm b}X^{\rm T}_{\rm s}A)}} = {\mbox{min}}\,{\frac{{\rm Tr}(A^{\rm T}S^{\rm s}_{\rm w}A)}{{\rm Tr}(A^{\rm T}S^{\rm s}_{\rm b}A)}} \enspace ,  \label{eq:graph_embed_source} \end{equation} \end{normalsize}
      \begin{normalsize} \begin{equation} {\mbox{min}}\,{\frac{{\rm Tr}(B^{\rm T}X_{\rm u}L^{\rm u}_{\rm w}X^{\rm T}_{\rm u}B)}{{\rm Tr}(B^{\rm T}X_{\rm u}L^{\rm u}_{\rm b}X^{\rm T}_{\rm u}B)}} = {\mbox{min}}\,{\frac{{\rm Tr}(B^{\rm T}S^{\rm u}_{\rm w}B)}{{\rm Tr}(B^{\rm T}S^{\rm u}_{\rm b}B)}} {\enspace ,}  \label{eq:graph_embed_target} \end{equation} \end{normalsize}
      where
       \[\begin{aligned} S^{\rm s}_{\rm b} & = X_{\rm s}L^{\rm s}_{\rm b}X^{\rm T}_{\rm s} \enspace ,\quad  S^{\rm s}_{\rm w}  =  X_{\rm s}L^{\rm s}_{\rm w}X^{\rm T}_{\rm s} \enspace , \\
       S^{\rm u}_{\rm b} & =  X_{\rm u}L^{\rm u}_{\rm b}X^{\rm T}_{\rm u} \enspace , \quad  S^{\rm u}_{\rm w}  =  X_{\rm u}L^{\rm u}_{\rm w}X^{\rm T}_{\rm u} \enspace , \end{aligned}\]
      where $L^{\rm s}_{\rm w}$ and $L^{\rm s}_{\rm b}$ are the Laplacian matrices of the intrinsic graph and the penalty graph for the source domain~\cite{yan2007graph,li2017low}. Similarly, $L^{\rm u}_{\rm w}$ and $L^{\rm u}_{\rm b}$ are the Laplacian matrices for the target domain. The Laplacian graphs guarantee that samples from the same category can be close to each other, and at the same time, samples from different categories can be separable. More details of the Laplacian graphs can be found in previous work~\cite{yan2007graph,li2017low}. Mathematically, $L = D - W$, $D$ is a diagonal matrix and its diagonal entry is ${D}_{ii}=\sum_{j\neq i}{W}_{ij}$. 
       $W_{\rm w}$ and $W_{\rm b}$ are the weight matrices for the intrinsic graph and the penalty graph, respectively. In this paper, we deploy the following two criteria to construct $W_{\rm w}$ and $W_{\rm b}$:
      \begin{enumerate}[(a)]
       \item Construct the intrinsic weight matrix $W_{\rm w}$: For each sample $x$, connect the nearest neighbor pair $v$ and $x$ if $v$ has the same label information with $x$.
      \item Construct the penalty weight matrix $W_{\rm b}$: For each domain, connect the $k$-nearest vertex pairs where samples in each pair belong to different classes.
      \end{enumerate}

      By applying (a), samples from the same class can be more compact and the local manifold structure can be preserved. By deploying (b), samples from the same domain but different classes can be more separable and the global discriminative information can be retained.
      We apply the heat kernel function to calculate $W_{\rm w}$ and $W_{\rm b}$, i.e., if two samples $x_i$ and $x_j$ are connected, then the weight between them would be ${\rm exp}(- (||x_i - x_j||^2)/2)$, otherwise it would be $0$.\\

      \subsubsection{Overall Objective Function}
        Considering all the above discussions, we have the overall objective function:
        \begin{normalsize} \begin{equation}  \underset{A,B}{\mbox{min}}\;{\frac 
        {{\rm Tr}\left(\begin{bmatrix}A^{\rm T} & B^{\rm T}\end{bmatrix}{\begin{bmatrix}M_{\rm ss}+\gamma S^{\rm s}_{\rm w} & M_{\rm su}\\ M_{\rm us} & M_{\rm uu}+\gamma S^{\rm u}_{\rm w}+ \mu I \end{bmatrix}}{\begin{bmatrix}A\\B\end{bmatrix}}\right)}
        {{\rm Tr}\left({\begin{bmatrix}A^{\rm T} & B^{\rm T}\end{bmatrix}}{\begin{bmatrix}\gamma S^{\rm s}_{\rm b} & {\bf 0}\\{\bf 0} &\gamma S^{\rm u}_{\rm b}+\mu S^{\rm u}_{\rm h}\end{bmatrix}}{\begin{bmatrix}A\\B\end{bmatrix}}\right)}} \enspace ,   \label{eq:obj_func} \end{equation} \end{normalsize}
        where \begin{equation}S^{\rm u}_{\rm h} = X_{\rm u}(I_{\rm u} - \frac{1}{n_{\rm u}}{\bf 1}_{n_{\rm u}}{\bf 1}^{\rm T}_{n_{\rm u}}) X^{\rm T}_{\rm u} \label{eq:pca_term} \end{equation} is the covariance matrix of the target domain, $\gamma$ and $\mu$ are trade-off parameters for the locality preserving term and the target variance term, respectively.

  \begin{table*}[t!p]
    \begin{center}
    \begin{tabular}{|p{16.8cm}<{\raggedright}|}
      \hline
        { {\bf Algorithm~1: Locality Preserving Joint Transfer for Domain Adaptation}\;} \\
      \hline
        {\bf Input:} Source and target domain data: $X_{\rm s},X_{\rm l}$ and $X_{\rm u}$; \,labels for source domain data and a few target domain data:$y_{\rm s}$ and $y_{\rm l}$; \\~~~~~~~~~Parameters: $\delta = 0.5$, $d$, $\mu$, $\gamma$\\
        {\bf Output:} Predicted labels $y_{\rm u}$ for target domain unlabeled data \\
        {\bf begin} \\
            1: Initialize pseudo labels of target domain unlabeled data $\hat y_{\rm u}$ using Label Propagation with $X_{\rm l}$, $X_{\rm u}$ and $y_{\rm l}$; \\~~~Compute $S^{\rm u}_{\rm h}$, $M_{\rm ss}$, $M_{\rm uu}$, $M_{\rm su}$, $M_{\rm us}$, $S^{\rm s}_{\rm b}$, $S^{\rm s}_{\rm w}$, $S^{\rm u}_{\rm b}$, $S^{\rm u}_{\rm w}$ according to (\ref{eq:pca_term})(\ref{eq:matrix_form_mss})(\ref{eq:graph_embed_source})(\ref{eq:graph_embed_target}), respectively; \\
        ~~{\bf while} not converge {\bf do}\\
            ~~~~~~2:    Solve the generalized eigen-decomposition problem in (\ref{eq:eigendecomp}) and select $d$ corresponding eigenvectors of $d$ \\~~~~~~~~ largest eigenvalues as the transformation $P$, and obtain transformation $A$ and $B$;\\
            ~~~~~~3:    Map the original data to respective subspace to get the embeddings: $Z_{\rm s} = A^{\rm T} X_{\rm s}$, $Z_{\rm l} = B^{\rm T} X_{\rm l}$,$Z_{\rm u} = B^{\rm T} X_{\rm u}$;\\
            ~~~~~~4:    Run Label Propagation on {$Z_{\rm s},Z_{\rm l},Z_{\rm u},y_{\rm s},y_{\rm l}$} to update pseudo labels in target domain $\hat y_{\rm u}$;\\
            ~~~~~~5:    Update landmark weights {$\alpha,\beta$} by (\ref{eq:qpsolver});\\
            ~~~~~~6:    Update $M_{\rm ss}, M_{\rm uu}, M_{\rm su}, M_{\rm us}, S^{\rm u}_{\rm b}, S^{\rm u}_{\rm w}$ by (\ref{eq:matrix_form_mss})(\ref{eq:graph_embed_source})(\ref{eq:graph_embed_target});\\
        ~~{\bf end while}\\
        {\bf end} \\
      \hline
    \end{tabular}
    \end{center}
    \vspace{-10pt}
    \end{table*}
  
    \subsection{Problem Optimization}
      \subsubsection{Optimizing the space mappings $A$ and $B$}
        To optimize Eq.~(\ref{eq:obj_func}), we write $[A ; B]$ as $P$. Thus, the objective function can be rewritten as:
        \begin{normalsize} \begin{equation}  \underset{P}{\mbox{min}}\;{\frac{{\rm Tr}\left(P^{\rm T}{\begin{bmatrix}M_{\rm ss}+\gamma S^{\rm s}_{\rm w} &  M_{\rm su} \\ M_{\rm us} & M_{\rm uu}+\gamma S^{\rm u}_{\rm w}+\mu I\end{bmatrix}}P\right)}{{\rm Tr}\left(P^{\rm T}{\begin{bmatrix}\gamma S^{\rm s}_{\rm b} & {\bf 0}\\{\bf 0} &\gamma S^{\rm u}_{\rm b}+\mu S^{\rm u}_{\rm h}\end{bmatrix}}P\right)}} .  \label{eq:optim_ab}  \end{equation} \end{normalsize}

        We can reformulate Eq.~\eqref{eq:optim_ab} as:
        \begin{normalsize} \begin{equation} \begin{array}{c} \underset{P}{\mbox{max}}\;{\rm Tr}\left(P^{\rm T}{\begin{bmatrix}\gamma S^{\rm s}_{\rm b} &{\bf 0}\\{\bf 0} &\gamma S^{\rm u}_{\rm b}+\mu S^{\rm u}_{\rm h}\end{bmatrix}}P\right),  \\ {\mbox{s.t.}}~ {\rm Tr}\left(P^{\rm T}{\begin{bmatrix}M_{\rm ss}+\gamma S^{\rm s}_{\rm w} & M_{\rm su}\\ M_{\rm us} & M_{\rm uu}\!+\!\gamma S^{\rm u}_{\rm w}\!+\!\mu I\end{bmatrix}}P\right)\!\! =\! 1. \label{eq:matrix_ab}  \end{array} \end{equation} \end{normalsize}

        According to the constrained optimization theory, we introduce a Lagrange multiplier $\Phi$, and get the Lagrange function for Eq.~(\ref{eq:matrix_ab}) as follows: 
         \begin{equation} \begin{aligned} \mathcal{L} & = {{\rm Tr}\left(P^{\rm T}{\begin{bmatrix}\gamma S^{\rm s}_{\rm b} &{\bf 0}\\{\bf 0} &\gamma S^{\rm u}_{\rm b}+\mu S^{\rm u}_{\rm h}\end{bmatrix}}P\right)}  \\ & - \!{{\rm Tr}\!\left(\!\!\left(\!P^{\rm T}\!{\begin{bmatrix}M_{\rm ss}\!+\!\gamma S^{\rm s}_{\rm w} & M_{\rm su}\\ M_{\rm us} & M_{\rm uu}\!+\!\gamma S^{\rm u}_{\rm w}\!+\!\mu I\end{bmatrix}}P\! -\! I\!\right)\!\Phi\!\right)} \end{aligned}, \label{eq:lagrange} \end{equation} 
        where $\Phi = {\mbox {diag} }(\phi_1,\cdots,\phi_d)$ and $(\phi_1,\cdots,\phi_d)$ are the $d$ largest eigenvalues of the following eigendecomposition problem:
        \begin{small}\begin{equation}   {\begin{bmatrix}\gamma S^{\rm s}_{\rm b} &{\bf 0}\\{\bf 0} &\gamma S^{\rm u}_{\rm b}\!+\!\mu S^{\rm u}_{\rm h}\end{bmatrix}}\!P\! =\! {\begin{bmatrix}M_{\rm ss}\!+\!\gamma S^{\rm s}_{\rm w} & M_{\rm su} \\ M_{\rm us} & M_{\rm uu}\!+\!\gamma S^{\rm u}_{\rm w}\!+\! \mu I\end{bmatrix}}\!P\Phi. \label{eq:eigendecomp} \end{equation}\end{small}

        As a result, $P$ consists of the corresponding $d$ eigenvectors of Eq.~\eqref{eq:eigendecomp}. At last, the subspaces spanned by $A$ and $B$ can be obtained easily once the transformation matrix $P$ is obtained.

      \subsubsection{Optimizing sample weights $\alpha$ and $\beta$}

Regarding $A$ and $B$ as constants, the optimization of $\alpha$ and $\beta$ can be~written~as:
 \begin{equation}  \begin{array}{c}  \underset{\alpha,\beta}{\mbox{min}} \;{\frac{1}{2} \alpha^{\rm T} K_{\rm ss} \alpha - \alpha^{\rm T} K_{\rm su} \beta} \\ {\mbox{s.t.}}\;\{\alpha^c_i,\beta^c_i\}\in [0,1],{\frac{\alpha^{c^{\rm T}}{\bf 1}}{n^c_{\rm s}}} = {\frac{\beta^{c^{\rm T}}{\bf 1}}{n^c_{\rm u} }} = \delta \enspace , \label{eq:opti_alphabeta}  \end{array} \end{equation}  
      where $(K_{\rm ss})_{i,j}$ in $K_{\rm ss} \in {\mathbb R}^{n_{\rm s} \times n_{\rm s}}$ is the coefficient associated with $(A^{\rm T} x^i_{\rm s})^{\rm T} A^{\rm T} x^i_{\rm s}$, and $(K_{\rm su})_{i,j}$ in $K_{\rm su} \in {\mathbb R}^{n_{\rm s} \times n_{\rm u}}$ is the coefficient associated with $(A^{\rm T} x^i_{\rm s})^{\rm T} B^{\rm T} x^j_{\rm u}$.

      With the above formulation, we can apply Quadratic Programming (QP) solvers to optimize the equivalent problem:
      \begin{normalsize} \begin{equation} \underset{   \resizebox{.25\hsize}{!}{$ z_i \in [0,1],\\ Z^{\rm T} V = G $}   }{\mbox {min}}\quad {\frac{1}{2}} Z^{\rm T} B Z  \enspace , \label{eq:qpsolver} \end{equation} \end{normalsize}
      where
      \begin{normalsize} 
        \[  \begin{aligned}
         & Z  = \begin{pmatrix}\alpha \\ \beta\end{pmatrix} {\enspace ,}\;
         B  = \begin{pmatrix}K_{\rm ss} & -K_{\rm su} \\ -K_{\rm su}^{\rm T} & {\bf 0}\end{pmatrix}{\enspace ,}\; G  \in {\mathbb R}^{1 \times 2C} \; {\mbox{with}} \\
          & (G)_c  = \begin{cases}
                  \delta n^c_{\rm s} & \mbox{if c $\leq$ C} \\ \delta n^{c-C}_{\rm u} & \mbox{if c $>$ C}
                \end{cases}  {\enspace ,} \\
           & V  = \begin{bmatrix}V_{\rm s} & {\bf 0}_{n_{\rm s}\times C} \\ {\bf 0}_{n_{\rm u} \times C} & V_{\rm u}\end{bmatrix} \in {\mathbb R}^{(n_{\rm s}+n_{\rm u})\times 2C} \; {\mbox{with}} \\ 
           & (V_{\rm s})_{ij}  = \begin{cases}
                      1 & \mbox{if $x^i_{\rm s} \in$ class $j$} \\
                        0 & \mbox{otherwise}
                  \end{cases} {\enspace ,~} {\mbox{and}} \\
         &(V_{\rm u})_{ij}  = \begin{cases}
                      1 & \mbox{if $x^i_{\rm u}$ predicted as class $j$} \\ 
                      0 & \mbox{otherwise}
                 \end{cases} {\enspace .}\\
        \end{aligned}  \] 
      \end{normalsize}

      At last, we use the label propagation~\cite{zhu2002learning} to train a classifier on the learned space and get the labels of the target domain. We show the procedures of our method in {Algorithm~1}.

\subsection{Computational Complexity}
     Now we analyze the computational complexity of Algorithm~1 by the big O notation. As stated above, $X_{\rm s} \in {\mathbb R}^{d_s \times n_{\rm s}},\, X_{\rm t} \in {\mathbb R}^{d_t \times n_{\rm t}},\,X = [X_{\rm s};X_{\rm t}] \in {\mathbb R}^{m \times n},\, A \in {\mathbb R}^{d_s \times d},\, B \in {\mathbb R}^{d_t \times d}$, where $d_s$ and $d_t$ are the original dimensionalities, $d$ is the dimensionality of the subspace, $n = n_{\rm s} + n_{\rm t}$ is the number of all samples. We denote the number of classes as $C$ and the number of iterations as $T$. The time cost of Algorithm~1 consists of the following three parts:
     \begin{enumerate}[1)]
      \item Solving the eigendecomposition problem in step 2 costs $O(Tmd^2)$.
      \item Solving the QP problem in step 5 consumes a maximum of $O(Tn^3)$.
      \item Computing the MMD matrices and the graph embedding matrices in step 6 costs $O(TCn^2)$.
     \end{enumerate}

     Then, the overall computational complexity of Algorithm 1 is $O(Tdm^2 + Tn^3 + TCn^2)$. In section IV, we will show that the number of iterations $T$ is usually smaller than $10$, which is enough to guarantee convergence. Besides, the typical values of $d$ are not greater than $200$, so $T \ll \mbox {min}(m,n) $, $d \ll \mbox {min}(m,n) $. Therefore, Algorithm~1 can be solved in polynomial time with respect to the number of samples.

\section{Experiments}


\subsection{Datasets Description}
{\bf Office+Caltech} is an extension of Office benchmark introduced in~\cite{saenko2010adapting}. Office includes object categories from Amazon ({\bf A}, images downloaded from amazon.com), DSLR ({\bf D}, high-resolution images by a digital SLR camera) and Webcam ({\bf W}, low-resolution images by a web camera). A, D and W are three different domains, and each domain consists of $31$ categories, e.g., monitor, keyboard and laptop.~4,652 images are contained in Office. Office+Caltech contains an additional domain, Caltech-256 ({\bf C}) \cite{griffin2007caltech}, which has~30,607 images and~256 categories. Then,~10 common classes to all four datasets are selected. There are~8 to~151 samples per category per domain, and~2,533 images in total. Furthermore, 800-dimensional SURF features, 4,096-dimensional DeCAF$_6$ features~\cite{donahue2013decaf} and 4096-dimensional VGG-FC6 features~\cite{simonyan2014very} are extracted and further normalized to unit vectors.



\begin{table*}[t]
\begin{center}
\caption{Accuracy (\%) on the CMU PIE dataset.}
\vspace{-8pt}
\label{tab:pie}
\begin{tabu} to 0.78 \textwidth {llccccccc}
\Xhline{0.75pt}
 ${\rm Source}$ & ${\rm Target~~~}$  & ~~SVMt~~ & ~~GFK~\cite{gong2012geodesic} & ~~SA~\cite{fernando2013unsupervised} & ~~JDA~\cite{long2013transfer} & ~~TJM~\cite{long2014transfer} & ~JGSA~\cite{zhang2017joint} & LPJT [Ours]\\
\hline
\multirow{4}*{C05} & C07  & $33.52$  & $45.49$ & $35.54$ & $54.08$ & $34.87$ & $52.73$ & {\bf 80.11}\\
                           & C09 & $43.69$   & $50.31$ & $46.38$ & $60.54$ & $44.36$ & $51.84$ & {\bf72.49}\\
                           & C27 & $61.28$   & $65.82$ & $63.62$ & $85.97$ & $60.74$ & $73.72$& {\bf 95.31}\\
                           & C29 & $36.46$   & $41.97$ & $39.58$ & $49.33$ & $34.93$ & $52.39$& {\bf 64.64}\\                           
\hline 

\multirow{3}*{C07}       & C05 & $42.05$   & $46.91$ & $44.24$ & $60.41$ & $38.75$& $64.26$ &{\bf 80.34}\\
                           & C09 & $41.85$   & $56.74$ & $44.06$ & $55.45$ & $49.82$   & $58.88$ & {\bf 75.49}\\
                           & C27 & $65.64$   & $70.86$ & $66.45$ & $82.43$ & $63.41$   & $70.71$ & {\bf 94.86}\\
                           & C29 & $34.13$   & $41.85$ & $35.48   $ &$47.37$& $35.23$ & $49.02$ &  {\bf69.73}\\
\hline 
\multirow{3}*{C09}       & C05  & $49.58$   & $48.65$ & $51.80$ & $58.64$ &$43.04$ & $64.89$ & {\bf 78.75}\\
                           & C07  & $42.91$   & $56.23$ & $46.41$ & $49.60$& $38.74$ & $59.91$ & {\bf 79.19}\\
                           & C27  & $67.98$   & $74.95$ & $68.76$ & $67.80$ & $65.33$& $72.63$ & {\bf 96.00}\\
                           & C29  & $42.40$   & $50.61$ & $45.22$ & $46.69$ & $41.12$& $57.72$ & {\bf 74.69}\\
\hline 
\multirow{3}*{C27}       & C05  & $66.96$   & $71.43$ & $69.57$ & $79.26$ & $63.39$& $74.73$ & {\bf 95.98}\\
                           & C07  & $62.06$   & $81.34$ & $64.03$ & $77.59$& $60.59$ & $76.24$ & {\bf 96.13}\\
                           & C09  & $70.71$   & $86.34$ & $72.06$ & $77.45$& $74.39$ & $67.89$ & {\bf 94.24}\\
                           & C29  & $54.23$   & $59.87$ & $55.94$ & $58.88$& $50.61$ & $63.05$ & {\bf 84.74}\\
\hline 
\multirow{3}*{C29}       & C05  & $46.16$   & $39.98$ & $48.50$ & $51.02$ & $37.18$& $63.99$ & {\bf 77.46}\\
                           & C07  & $34.81$   & $38.80$ & $35.30$ & $42.79$& $29.28$ & $54.02$ & {\bf 69.06}\\
                           & C09  & $47.98$   & $48.96$ & $49.39$ & $41.85$& $42.77$ & $59.87$ & {\bf 77.63}\\
                           & C27  & $59.12$   & $54.73$ & $59.51$ & $62.51$& $46.77$ & $66.39$ & {\bf 90.36}\\                           
\hline
\multicolumn{2}{c}{${\rm Avg.}$}  & $50.17$ & $56.59$ & $52.09$& 60.48 & $47.77$ & $62.74$ & ${\bf 82.36}$\\
\Xhline{0.75pt}
\end{tabu}
\end{center}
\vspace{-10pt}
\end{table*}

\begin{table*}[]
\begin{center}
\caption{Accuracy (\%) on the USPS and MNIST datasets.}
\vspace{-8pt}
\label{tab:digits}
\begin{tabu} to 0.78 \textwidth {llcccccccc}
\Xhline{0.75pt}
${\rm Source~}$ & ${\rm Target~}$  & SVMt   & SA~\cite{fernando2013unsupervised}    & SDA~\cite{Kan_2015_ICCV}   & GFK~\cite{gong2012geodesic}  & TCA~\cite{pan2011domain}   & JDA~\cite{long2013transfer}   & TJM~\cite{long2014transfer}     & LPJT [Ours]  \\
\hline
MNIST & USPS & 65.94 & 67.78 & 65.00 & 61.22 & 56.33 & 67.28 & 63.28  & {\bf 73.33} \\
USPS & MNIST & 44.70 & 48.80 & 35.70 & 46.45 & 51.20 & 59.65 & 52.25  & {\bf 63.60} \\
\hline
\multicolumn{2}{c}{${\rm Avg.}$}  & 55.32 & 58.29 & 50.35 & 56.84 & 53.77 & 63.47 & 57.77  & {\bf 68.47}  \\ 
\Xhline{0.75pt}
\end{tabu}
\end{center}
\vspace{-10pt}
\end{table*}

\begin{table*}[t]
\begin{center}
\caption{Domain adaptation results (\%) on Office+Caltech dataset with DeCAF$_6$ features.}
\vspace{-8pt}
\label{tab:decaf2decaf}
\begin{tabu} to 0.8\textwidth {llccccccc}
\Xhline{0.75pt}
Source~~~~            & Target~~~  & ~~SVMt~~    & PCA+SVM  & SHFA~\cite{li2014learning}  & CDLS~\cite{hubert2016learning}  & DDC~\cite{tzeng2014deep} & DAN~\cite{long2015learning} & LPJT [Ours]  \\ 
\hline
\multirow{2}{*}{Caltech} 
                         & Amazon  & 88.14 & 43.01   & 89.18 & 89.22 & 91.90 & 92.01 & {\bf 92.06} \\ 
                         & Webcam  & 88.64 & 29.43   & 90.26 & 91.70 & 85.40 & 90.62 & {\bf 92.72} \\ \hline
\multirow{2}{*}{Amazon}  & Caltech & 80.02 & 22.93   & 81.99 & 84.72 & 85.03 & 85.12 & {\bf 85.44} \\ 
                         & Webcam  & 88.64 & 27.21   & 89.74 & 92.08 & 86.11 & {\bf 93.83} & 92.23 \\ \hline
\multirow{2}{*}{Webcam}  & Caltech & 80.02 & 20.84   & 82.14 & 84.81 & 78.02 & 84.33 & {\bf 86.00} \\ 
                         & Amazon  & 88.14 & 36.09   & 89.66 & 88.90 & 84.91 & 92.12 & {\bf 92.33} \\ 
                         \hline
\multicolumn{2}{c}{Avg.}         & 85.60    & 29.92 & 87.16 & 88.57 & 85.23 & 89.67 & ~~{\bf 91.79}~~ \\ 
\Xhline{0.75pt}
\end{tabu}
\end{center}
\vspace{-10pt}
\end{table*}

{\bf CMU PIE}~\cite{cmu_pie} includes 41,368 images of 68 person with different poses, different illumination conditions, and different expressions. We choose the most widely used five poses distributed by Cai et al.~\cite{cai2007spectral}, i.e., C05, C07, C09, C27 and C29, for our experiments. All face images were converted to grayscale, cropped, and resized to $32 \times 32$ pixels. 

{\bf USPS} and {\bf MNIST} are two datasets of handwritten digits. USPS contains 9,298 handwritten digit images of $16 \times 16$ pixel values in total, which is split into 7,291 training images and 2,007 test images. The MNIST dataset has a training set of 60,000 examples, and a test set of 10,000 examples. Both the training set and test set are used in our experiments. 

{\bf Office-Home}~\cite{venkateswara2017Deep} consists of 4 subsets: Art (Ar), Product (Pr), Real-world (Rw) and Clipart (Cl). The images in Art category are paintings, sketches and artistic depictions. Images in the Product category are collected without background. Real-word images are regular images captured with a camera. Each subset is regarded as a domain. In total, there are 65 object categories and 15,500 images in the dataset. In this paper, we follow previous work~\cite{venkateswara2017Deep} and extract deep features of fc7 layer with a VGG-F model, which is pre-trained on ImageNet 2012.

{\bf VisDA} 2017 challenge (VisDA2017)~\cite{visda2017} is a large-scale dataset with over 280K images across 12 categories. The dataset contains a training domain which is synthetic 2D renderings of 3D models generated from different angles and with different lighting conditions, a photo-realistic or real-image validation domain, and a test domain which is different from the validation domain and without labels. In this paper, we use the training domain as the source domain and the validation domain as the target domain.


\subsection{Protocols}

Our approach can handle both homogeneous and heterogeneous domain adaptation problems. In this section, we first report the homogeneous domain adaptation results with unsupervised settings. Then, we report the results of heterogeneous domain adaptation. For the tested datasets, each subset is regarded as a domain. By picking any two to build a cross-domain task, we can generate $3\times 4=12$ evaluations for both Office+Caltech and Office-Home datasets,  and $4\times 5=20$ evaluations for the CMU PIE dataset. We further normalize all of the data using z-score. For the parameters, we fix the average weight $\delta = 0.5$. Both of the numbers of neighbors in the intrinsic graph and the penalty graph are $5$. The number of iterations $T$ is set to $5$. For different experimental tasks, we set different values for the hyper-parameters to gain good performance. Specifically, when evaluating our method on Office+Caltech with ${\rm SURF}$ features, we set $\mu = 0.1$, $\gamma = 0.001$ and $d = 40$. For Office+Caltech with ${\rm DeCAF_6}$ features~\cite{donahue2013decaf}, we set $\mu = 0.5$, $\gamma = 0.01$ and $d = 40$. For CMU PIE dataset, we set $\mu = 0.1$, $\gamma = 0.02$ and $d = 120$. For Office-Home dataset, we set $\mu = 1$, $\gamma = 0.05$ and $d = 200$. For VisDA dataset, we set $\mu = 0.1$, $\gamma = 0.05$ and $d = 30$. Since there are no labeled target samples in unsupervised domain adaptation, we follow previous work~\cite{gong2012geodesic,long2013transfer} and set the hyper-parameters by searching a wide range and use the optimal ones. In semi-supervised domain adaptation, one can deploy cross-validation to tune them.

We follow the same experimental protocols in~\cite{gong2012geodesic,long2014transfer} to perform homogeneous domain adaptation. Both Office+Caltech and CMU PIE are tested. In homogeneous domain adaptation on CMU PIE, the projection matrix $A$ and $B$ have the same dimensionality, we encourage $A$ and $B$ to be close by minimizing $\|A-B\|_F^2$~\cite{jing2018learning}. Several state-of-the-art approaches, e.g., subspace alignment (SA)~\cite{fernando2013unsupervised}, geodesic flow kernel (GFK)~\cite{gong2012geodesic}, joint distribution analysis (JDA)~\cite{long2013transfer}, transfer joint matching (TJM)~\cite{long2014transfer}, joint geometrical and statistical alignment (JGSA)~\cite{zhang2017joint} are tested for comparison. In particular, SVM is also reported as a non-transfer baseline.

For heterogeneous domain adaptation, we follow the same settings in previous work~\cite{hubert2016learning,li2018transfer,li2014learning}. Specifically, for the source domains, $20$ samples per category are randomly selected as labeled data. For the target domain, $3$ samples per object are randomly selected as the gallery, and the rest images are used as probe data. Note that the gallery samples in
our experiments are only used for testing/classification, whereas they are also used for training in compared methods like CDLS. Four state-of-the-art methods are used for comparison, for instance,  joint geometrical and statistical alignment (JGSA)~\cite{zhang2017joint}, semi-supervised heterogeneous feature augmentation (SHFA)~\cite{li2014learning} and cross-domain landmark selection
(CDLS)~\cite{hubert2016learning}. In particular, PCA+SVM is reported as a baseline, and SVM is reported as a non-transfer baseline.

We also evaluate our method on large-scale dataset Office-Home and VisDA. The protocols of Office-Home are same with previous work~\cite{venkateswara2017Deep}. The protocols of VisDA are same with previous work~\cite{saito2018maximum}. In large-scale tests, we compare our method with not only traditional domain adaptation methods but also end-to-end deep approaches.

\subsection{Results and Discussions}
The results of homogeneous domain adaptation on CMU PIE are shown in Table~\ref{tab:pie}. It can be seen that our approach outperforms state-of-the-art methods on all of the $20$ evaluations. The average classification accuracy of our method is $82.36\%$, which overwhelmingly has a $19.62\%$ improvement compared with the best baseline JGSA. The results of homogeneous domain adaptation on MNIST and USPS datasets are shown in Table~\ref{tab:digits}. The results further verify the advances of our method. Compared with the MMD-based methods, e.g., TCA, JDA and JGSA, our method advances from two aspects. The first is that we pay attention to not only features but also samples. The other is that our method can preserve the local consistency of samples. Take CMU PIE as an example, the Euclidean distance between samples from one domain but different classes may be smaller than the distance of samples within one class but from different domains. Thus, neglecting the local consistency of samples will lead to misclassification. To handle this, our approach constructs an intrinsic graph to preserve the local manifold structure and a penalty graph to make samples with different labels more separable. This novel graph structure enhances the discriminability of the model, which was also verified in LRDE~\cite{li2017low}. 

\begin{table*}[t]
\caption{Heterogeneous domain adaptation results (\%) of SURF to VGG-FC6 on Office+Caltech dataset.}
\begin{center}
\vspace{-8pt}
\label{tab:surf2vgg}
\begin{tabu} to 0.78\textwidth {llcccccc}
\Xhline{0.75pt}
Source~~~~             & Target~~~~  & SVMt    & PCA+SVM & JGSA~\cite{zhang2017joint}  & SHFA~\cite{li2014learning}  & CDLS~\cite{hubert2016learning}  & LPJT [Ours]  \\ \hline
\multirow{3}{*}{Caltech} & Caltech & ~~73.33~~ & ~~28.41~~   & ~~79.99~~ & ~~75.30~~ & ~~77.31~~ & ~~{\bf 83.43}~~ \\ 
                         & Amazon  & 86.22 & 47.02   & 85.31 & 86.40 & 84.48 & {\bf 89.56} \\ 
                         & Webcam  & 89.43  & 42.08   & 83.74 & 91.21 & 89.06 & {\bf 92.45} \\ \hline
\multirow{3}{*}{Amazon}  & Caltech & 73.33 & 29.33   & 80.64 & 74.59 & 77.77 & {\bf 83.13} \\ 
                         & Amazon  & 86.22 & 35.40   & 86.94 & 86.82 & 84.70 & {\bf 90.22} \\ 
                         & Webcam  & 89.43  & 42.34   & 86.08 & 90.68 & 89.43 & {\bf 92.57} \\ \hline
\multirow{3}{*}{Webcam}  & Caltech & 73.33 & 29.24   & 80.89 & 75.27 & 77.31 & {\bf 82.76} \\ 
                         & Amazon  & 86.22 & 37.28   & 87.42 & 87.08 & 84.81 & {\bf 90.01} \\ 
                         & Webcam  & 89.43  & 41.70   & 86.98 & 90.23 & 89.81 & {\bf 92.60} \\ \hline 
\multicolumn{2}{c}{Avg.}         & 82.99   & 36.98   & 84.22 & 84.18 & 83.85 & {\bf 88.53} \\ \Xhline{0.75pt}
\end{tabu}
\end{center}
\vspace{-10pt}
\end{table*}


The results of homogeneous domain adaptation on Office+Caltech dataset with DeCAF$_6$ features are shown in Table~\ref{tab:decaf2decaf}. To verify the effectiveness of the proposed method, we also report the results of two end-to-end deep models, e.g., deep domain confusion (DDC)~\cite{tzeng2014deep} and deep adaption network (DAN)~\cite{long2015learning}. It is worth noting that dataset DSLR has the smallest number of samples, which may lead to over-fitting during training. Therefore, we did not report the results on DSLR. From the results in Table~\ref{tab:decaf2decaf}, we can see that our approach outperforms not only the traditional methods but also the end-to-end deep models, e.g., DDC and DAN. Specifically, our approach achieves 3.22\% improvements against CDLS and 2.12\% over DAN. In our experiments, we deploy DeCAF$_6$ features, DDC and DAN are based on AlexNet~\cite{krizhevsky2012imagenet} with MMD regularization. Actually, the basic techniques of our method and DDC/DAN are the same. Specifically, features are extracted by convolutional neural networks and knowledges are adapted by distribution matching. Compared with end-to-end deep models, our approach is more flexible. Different transfer learning techniques can be easily incorporated into it. 

It is worth noting that SVM$_t$ and PCA+SVM show a huge difference in the results. The reason is that SVM$_t$ directly classifies target samples with the labeled target samples as references. However, PCA+SVM classifies target samples with the labeled source samples as references. 

For heterogeneous domain adaptation, we evaluate our model on Office+Caltech dataset. DeCAF$_6$, VGG-FC6 and SURF features are tested. The $3$ different features can perform $3\times 2=6$ evaluations. However, SURF is a handcrafted feature, of which the representation ability is much weaker than deep features~\cite{li2018transfer,hubert2016learning,jing2018learning}. Therefore, it is meaningless to transfer knowledge from deep features to SURF. At the same time, we are interested in {\it if handcrafted features can assist deep features} and {\it if deep features can reinforce each other}. So we perform four evaluations on Office+Caltech dataset for heterogeneous domain adaptation.

The heterogeneous domain adaptation results of SURF to VGG-FC6 are reported in Table~\ref{tab:surf2vgg}. From the results, it is clear that the transfer learning algorithms perform better than baseline PCA+SVM and SVM$_t$. In other words, the results prove that the handcrafted features, e.g., SURF, can be used to reinforce the deep features. In general, deep features are automatically learned by an architecture with an explicit objective. However, handcrafted features, in most cases, are designed with human experience. For a specific task, the two have overlaps. For instance, human beings can recognize a face by observing the eyes, mouth and nose, which are also salient parts of a deep network. At the same time, human experiences do have differences with an automatic feature extractor. Some useful factors involved in the handcrafted features can complement the deep features. On the other hand, from the perspective of multi-view leaning, different kinds of features are different representations of the same information. Each of them has its pros and cons, joint optimizing them can strengthen its own strong points, and learn from the others.  

\begin{table*}[t]
\begin{center}
\caption{Heterogeneous domain adaptation results (\%) of DeCAF$_6$ to VGG-FC6 on Office+Caltech dataset.}
\label{tab:decaf2vgg}
\vspace{-8pt}
\begin{tabu} to 0.78\textwidth {llcccccc}
\Xhline{0.75pt}
Source~~~~             & Target~~~~  & SVMt    & PCA+SVM & JGSA~\cite{zhang2017joint}  & SHFA~\cite{li2014learning}  & CDLS~\cite{hubert2016learning}  & LPJT [Ours]  \\ \hline
\multirow{3}{*}{Caltech} & Caltech & ~~73.33~~ & ~~19.52~~   & ~~80.51~~ & ~~74.98~~ & ~~78.59~~ & ~~{\bf 82.72}~~ \\ 
                         & Amazon  & 86.22 & 39.11   & 88.90  & 87.03 & 85.02 & {\bf 90.42} \\ 
                         & Webcam  & 89.43  & 29.70   & 92.34 & 90.91 & 89.06 & {\bf 92.60} \\ \hline
\multirow{3}{*}{Amazon}  & Caltech & 73.33 & 27.87   & 80.2  & 75.41 & 78.87 & {\bf 82.59} \\ 
                         & Amazon  & 86.22 & 35.41   & 89.05 & 86.84 & 85.34 & {\bf 90.54} \\ 
                         & Webcam  & 89.43 & 26.49   & {\bf 92.72} & 90.53 & 89.81 & 92.60 \\ \hline
\multirow{3}{*}{Webcam}  & Caltech & 73.33 & 14.39   & 79.51 & 75.4  & 79.23 & {\bf 80.96} \\ 
                         & Amazon  & 86.22 & 31.63   & 89.69 & 87.1  & 85.24 & {\bf 90.88} \\ 
                         & Webcam  & 89.43  & 32.34   & {\bf 92.98} & 90.57 & 89.43 & 92.57 \\  \hline
\multicolumn{2}{c}{Avg.}         & 82.99   & 28.49   & 87.32 & 84.31 & 84.51 & {\bf 88.43} \\ \Xhline{0.75pt}
\end{tabu}
\end{center}
\vspace{-10pt}
\end{table*}

\begin{table*}[t]
\begin{center}
\caption{Heterogeneous domain adaptation results (\%) of VGG-FC6 to DeCAF$_6$ on Office+Caltech dataset.}
\label{tab:vgg2decaf}
\vspace{-8pt}
\begin{tabu} to 0.78\textwidth {llcccccc}
\Xhline{0.75pt}
Source~~~~            & Target~~~~  & SVMt    & PCA+SVM & JGSA~\cite{zhang2017joint}  & SHFA~\cite{li2014learning}  & CDLS~\cite{hubert2016learning}  & LPJT [Ours]  \\ \hline
\multirow{3}{*}{Caltech} & Caltech & ~~80.02~~ & ~~25.10~~   & ~~84.67~~ & ~~81.79~~ & ~~83.99~~ & ~~{\bf 85.25}~~ \\ 
                         & Amazon  & 88.14 & 31.64   & 90.2  & 89.25 & 88.90 & {\bf 91.84} \\ 
                         & Webcam  & 88.64 & 33.58   & {\bf 92.87} & 89.85 & 92.08 & 92.72 \\ \hline
\multirow{3}{*}{Amazon}  & Caltech & 80.02 & 20.30   & 84.63 & 81.96 & 84.45 & {\bf 86.21} \\ 
                         & Amazon  & 88.14 & 29.80   & 90.54 & 89.52 & 88.79 & {\bf 92.13} \\ 
                         & Webcam  & 88.64 & 21.58   & {\bf 93.43} & 89.55 & 91.70 & 92.72 \\ \hline
\multirow{3}{*}{Webcam}  & Caltech & 80.02 & 14.13   & 83.68 & 82.08 & 84.26 & {\bf 86.08} \\ 
                         & Amazon  & 88.14 & 29.38   & 90.44 & 89.66 & 89.33 & {\bf 92.10} \\ 
                         & Webcam  & 88.64 & 28.83   & {\bf 93.74} & 90.00 & 92.08 & 92.72 \\ \hline 
\multicolumn{2}{c}{Avg.}         & 85.60    & 26.04   & 89.36 & 87.07 & 88.4  & {\bf 90.2}  \\ \Xhline{0.75pt}
\end{tabu}
\end{center}
\vspace{-10pt}
\end{table*}

\begin{figure*}[t]
\begin{center}
 \includegraphics[width=0.7\linewidth]{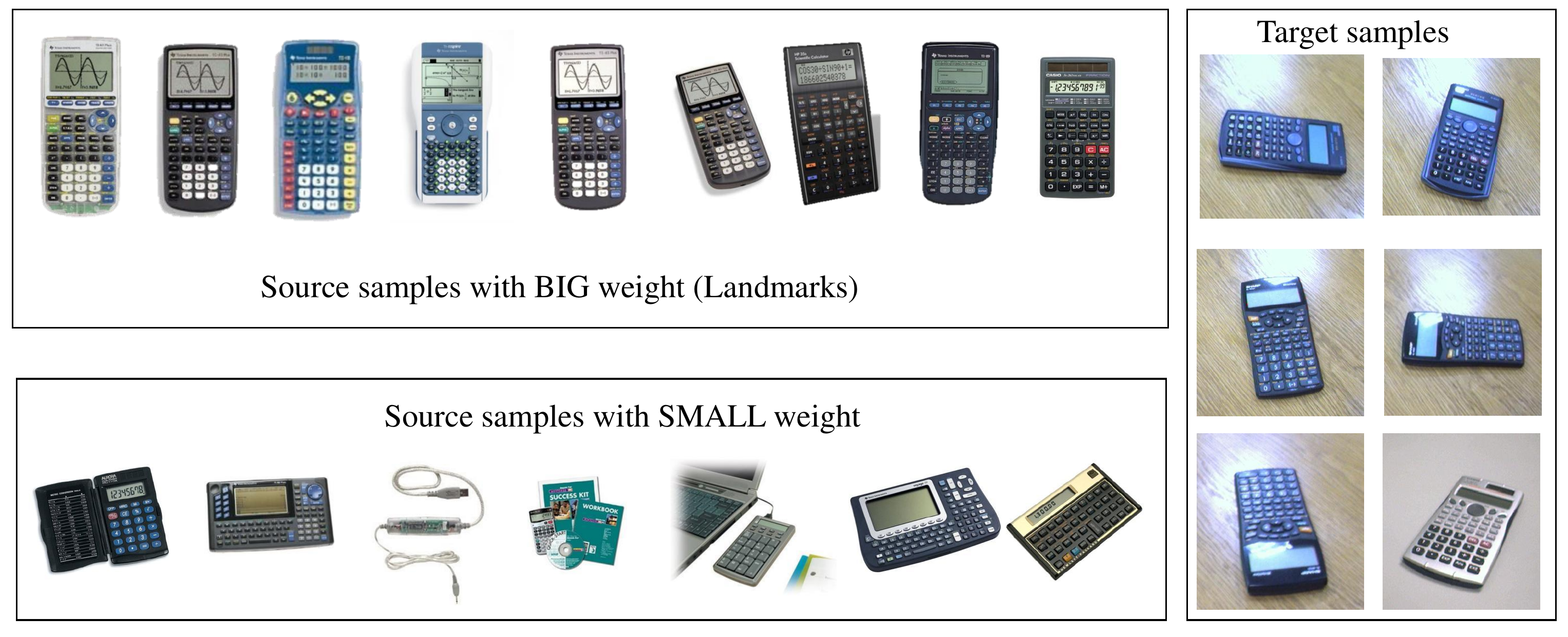}
 \vspace{-10pt}
\caption{The visual results of landmark selection on Amazon$\rightarrow$ Webcam. Samples with the label ``calculator'' are used as examples. It can be seen that the selected landmarks are more similar to target samples.}
\label{fig:landmarks}
\end{center}
\vspace{-15pt}
\end{figure*}

Generally, transfer learning is regarded as a technique of leveraging knowledge from one domain to another different domain. The experiments in this work, especially the adaptations on the same dataset, e.g., Caltech $\rightarrow$ Caltech, give us a clue that we can also perform intra-domain knowledge transfer among different feature types. This conclusion is also verified by the results in Table~\ref{tab:decaf2vgg} and Table~\ref{tab:vgg2decaf} which report the heterogeneous domain adaptation of DeCAF$_6$ to VGG-FC6 and VGG-FC6 to DeCAF$_6$. The results demonstrate that transfer learning can be used to complement one deep feature by another one. The results also show that our approach is more effective than previous works.

\begin{figure*}[t]
\begin{center}
\subfigure[Results with varying $\gamma$;]{
 \includegraphics[width=0.23\linewidth]{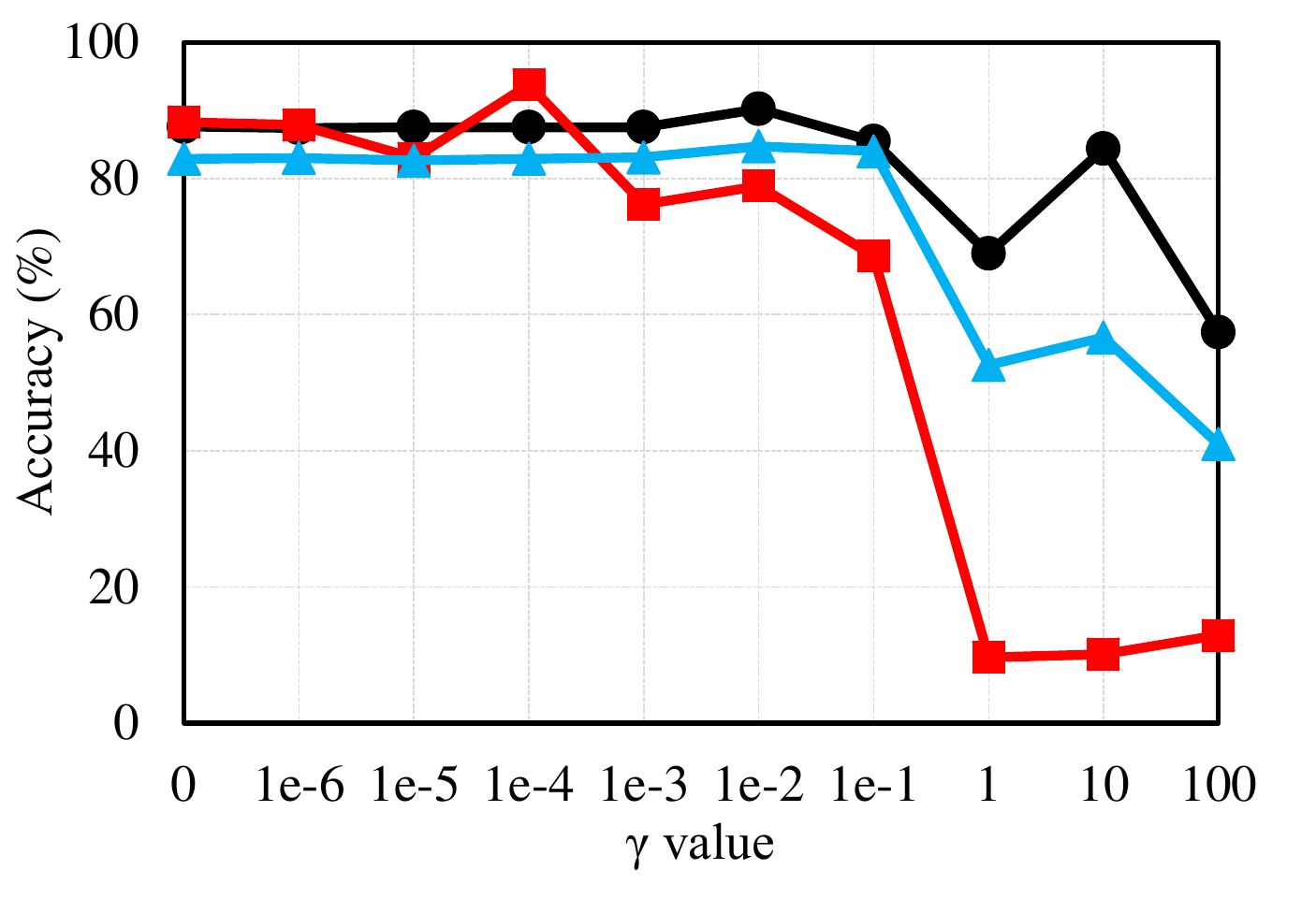}
}
\subfigure[Results with varying $\mu$;]{
  \includegraphics[width=0.23\linewidth]{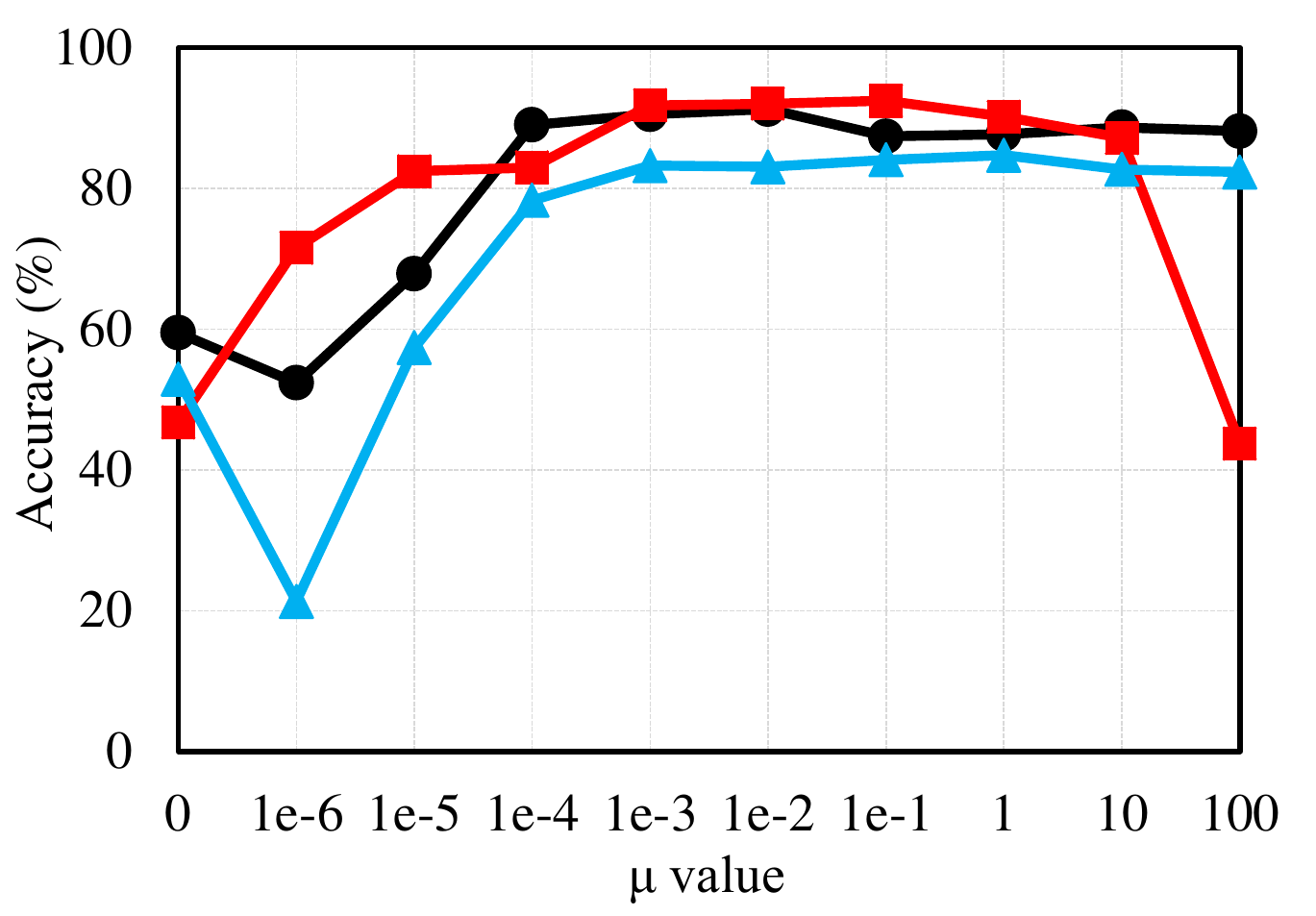}
}
\subfigure[Results with varying dimensions;]{
 \includegraphics[width=0.23\linewidth]{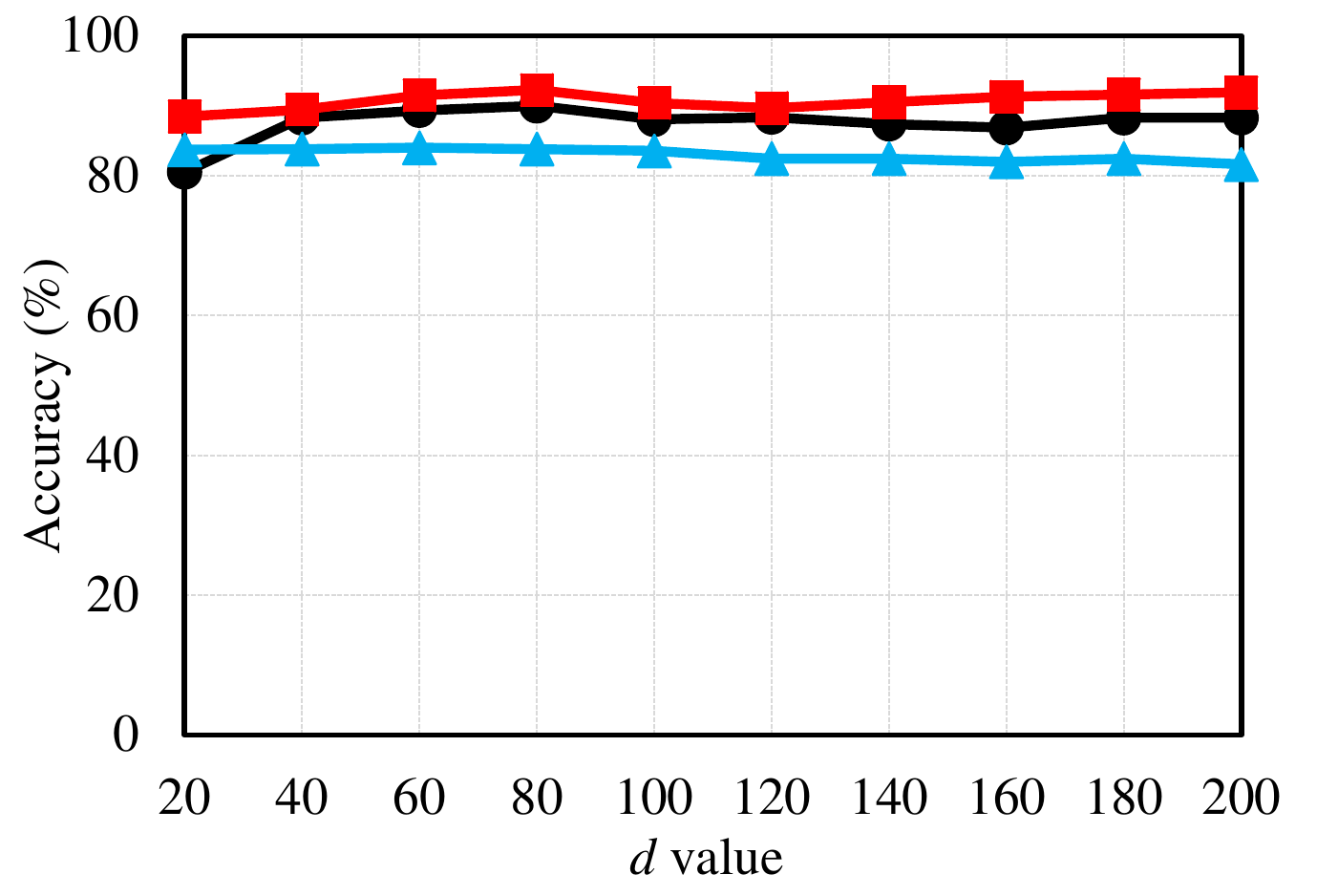}
}
\subfigure{
 \includegraphics[width=0.5\linewidth]{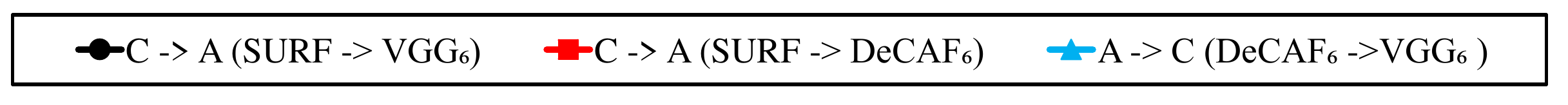}
}
\vspace{-10pt}
\caption{Parameter sensitivity. The results suggest that $\gamma$ should be set as a small value ($\ll 1$), $\mu$ is optimal in the range of [$10^{-3}$,1], and $d$ is not sensitive.}
\label{fig:paras}
\end{center}
\vspace{-15pt}
\end{figure*}

\begin{figure}[t]
\begin{center}
\subfigure[Objective value with iterations;]{
 \includegraphics[width=0.42\linewidth]{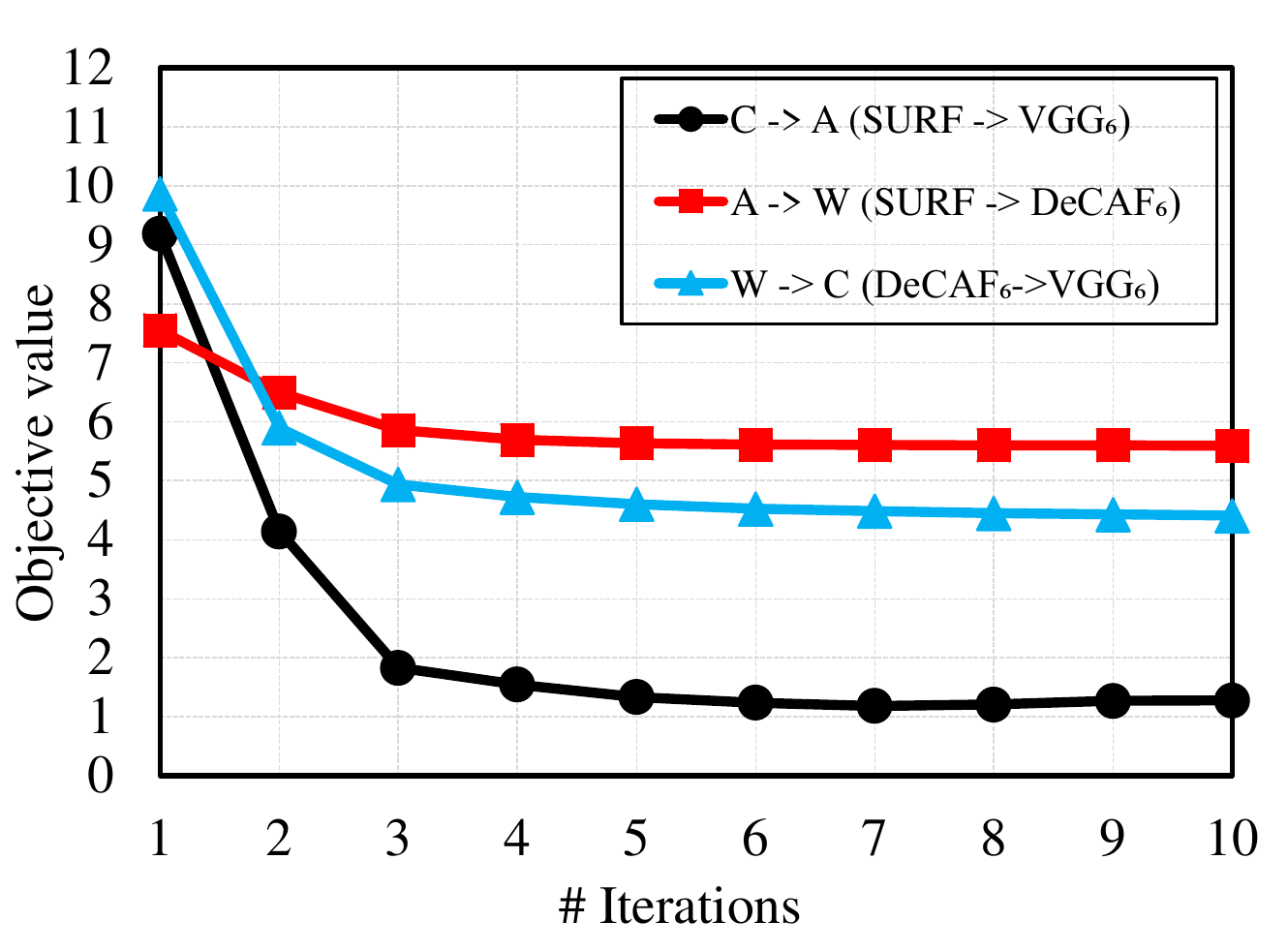}
}
\subfigure[MMD distance with iterations;]{
  \includegraphics[width=0.43\linewidth]{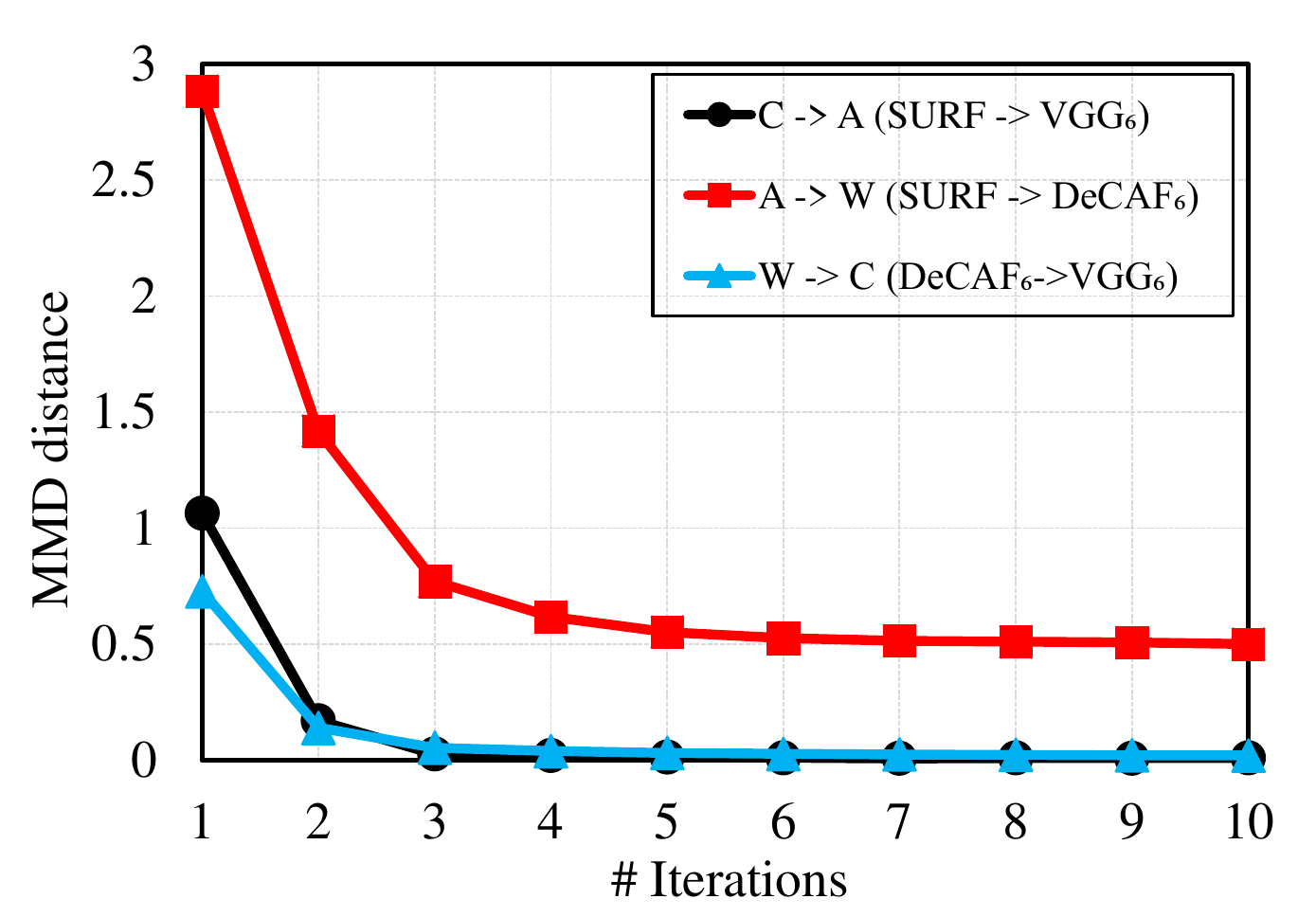}
}
\vspace{-5pt}
\caption{Convergence analysis. It can be seen that our method is able to achieve steady results after 5 iterations. The domain discrepancy (MMD distance) can be reduced by our method.}
\label{fig:convergance}
\end{center}
\vspace{-16pt}
\end{figure}

\subsection{Large-scale Evaluations}
The previous experiments have demonstrated the effectiveness of our method on standard datasets. In this section, we further evaluate our algorithm on two large-scale datasets. Specifically, we compare our method with several state-of-the-art deep methods, e.g., MCD~\cite{saito2018maximum}, SimNet~\cite{pinheiro2018unsupervised} and DHN~\cite{venkateswara2017Deep}.

The results on Office-Home dataset and VisDA dataset are reported in Table~\ref{tab:office-home} and Table~\ref{tab:visda2}, respectively. It can be seen that our method performs better than not only traditional methods but also deep methods. For instance, we achieved~2\% accuracy improvement against DHN~\cite{venkateswara2017Deep} on office-home dataset and MCD~\cite{saito2018maximum} on VisDA dataset. Both DHN and MCD are based on state-of-the-art deep neural networks. Compared with traditional methods, deep methods integrate feature extraction and knowledge transfer into a unified network and achieve promising results. However, some techniques, which have been proven effective in domain adaptation, are hard to be implemented with deep structure. For example, the conditional MMD can be easily optimized by matrix operations but it is very tricky in deep networks. In addition, the results on Office-Home and VisDA verify that our method is applicable to large-scale dataset and is able to achieve favorable accuracy. Furthermore, our method generally runs faster than deep ones since we use off-the-shelf features. 

\begin{table*}[t]
\begin{center}
\caption{Accuracy (\%) on the Office-Home dataset.}
\label{tab:office-home}
\vspace{-8pt}
\begin{tabu} to 0.9 \textwidth {llcccccccccc}
\Xhline{0.75pt}
 ${\rm Source}$ & ${\rm Target}$  & \!\!GFK\cite{gong2012geodesic}\!\! & \!\!JDA\cite{long2013transfer}\!\! & \!\!LSC\cite{hou2016unsupervised}\!\! & \!\!\!RTML\cite{ding2017robust}\!\!\! & \!\!\!JGSA\cite{zhang2017joint}\!\!\! & \!\!\!\!pUnDA\cite{gholami2017punda}\!\!\!\! & \!\!DAN\cite{long2015learning}\!\! & \!\!\!DHN\cite{venkateswara2017Deep}\!\!\! & \!\!\!WDAN\cite{yan2017mind}\!\!\! & \!\!LPJT[Ours]\!\!\\
\hline
\multirow{3}*{Art} & Clipart   & $21.60$   & $25.34$ & $31.81$ & $27.57$ & $28.81$ & $29.99$ & $30.66$ & $31.64$ & $32.26$ &{\bf 32.53}\\
                   & Product   & $31.72$   & $35.98$ & $39.42$ & $36.20$ & $37.57$ & $37.76$ & $42.17$ & $40.75$ & $43.16$ & {\bf 54.83}\\
                   & RelWld & $38.83$   & $42.94$ & $50.25$ & $46.09$ & $48.92$ & $50.17$ & $54.13$ & $51.73$ & $54.98$ & {\bf 57.13}\\
                                                    
\hline 

\multirow{3}*{\!\!Clipart\!\!}   & Art         & $21.63$   & $24.52$ & $35.46$ & $29.49$ & $31.67$ & $33.90$ & $32.83$ & $34.69$ & $34.28$ & {\bf 34.44}\\
                           & Product   & $34.94$   & $40.19$ & $51.19$ & $44.69$ & $46.30$ & $48.91$ & $47.59$ & $51.93$ & $49.92$ & {\bf 53.77}\\
                           & RelWld & $34.20$   & $40.90$ & $51.43$ & $44.66$ & $46.76$ & $48.71$ & $49.78$ & $52.79$ & $50.26$ & {\bf 53.00}\\                          
\hline 
\multirow{3}*{\!\!Product\!\!}     & Art        & $24.52$   & $25.96$ & $30.46$ & $28.21$ & $28.72$ & $30.31$ & $29.07$ & $29.91$ & $30.82$ & {\bf 35.60}\\
                           & Clipart    & $25.73$   & $32.72$ & $39.54$ & $36.12$ & $35.90$ & $38.69$ & $34.05$ & {\bf 39.63} & $38.27$ & 35.28\\
                           & RelWld  & $42.92$   & $49.25$ & $59.74$ & $52.99$ & $54.47$ & $56.91$ & $56.70$ & $60.71$ & $56.87$ & {\bf 60.94}\\
                           
\hline 
\multirow{3}*{\!\!RealWld\!\!}   & Art        & $32.88$   & $35.10$ & $43.98$ & $38.54$ & $40.61$ & $42.25$ & $43.58$ & $44.99$ & $44.32$ & {\bf 45.58}\\
                           & Clipart    & $28.96$   & $35.35$ & $42.88$ & $40.62$ & $40.83$ & $44.51$ & $38.25$ & {\bf 45.13} & $39.35$ & 39.43\\
                           & Product  & $50.89$   & $55.35$ & $62.25$ & $57.80$ & $59.16$ & $61.05$ & $62.73$ & $62.54$ & $63.34$ & {\bf 67.81}\\
                                                                               
\hline
\multicolumn{2}{c}{${\rm Avg.}$}  & $32.40$   & $36.97$ & $44.87$ & $40.25$ & $41.64$ & $43.59$ & $43.46$ & $45.53$ & $44.81$ & {\bf 47.52}\\
\Xhline{0.75pt}
\end{tabu}
\end{center}
\vspace{-10pt}
\end{table*}

\begin{table*}[ht!p]
\centering
\caption{Class-wise results (accuracy~\%) of on VisDA dataset. }
\label{tab:visda2}
\vspace{-8pt}
\begin{tabular}{lccccccccccccc}
\Xhline{0.75pt}
 Methods & ~aero.~ & ~bike~ & ~~bus~~ & ~~car~~ & horse & knife & moto. & person & plant & sktbrd & train & truck & ~~Mean~~\\
\hline
GFK ~\cite{gong2012geodesic} & 77.8 & 58.7 & 60.7 & 42.0 & 88.3 & 44.2 & 80.0 & 60.3 & 72.5 & 42.1 & 54.2 & 19.3 & 58.3 \\

TCA ~\cite{pan2011domain} & 87.1 & 65.8 & 62.5 & 51.4 & 92.3 & 28.7 & 78.2 & 77.8 & 86.5 & 36.2 & 72.0 & 27.8 & 63.9 \\

JDA ~\cite{long2013transfer} & 85.3 & 46.5 & 55.0 & 56.3 & 88.7 & 45.1 & 85.4 & 77.5 & 68.0 & 33.5 & 70.1 & 19.3 & 60.9 \\

JGSA ~\cite{zhang2017joint} & 91.1 & 63.8 & 64.2 & 47.5 & 92.6 & 42.3 & 79.8 & 76.5 & 84.3 & 50.1 & 72.2 & 26.5 & 65.8 \\
\hline 
RestNet ~\cite{he2016deep} & 55.1 & 53.3 & 61.9 & 59.1 & 80.6 & 17.9 & 79.7 & 31.2 & 81.0 & 26.5 & 73.5 & 8.5 & 52.4\\
DANN \cite{ganin2014unsupervised}~~~~~~~ & 81.9 & 77.7 & {82.8} & 44.3 & 81.2 & 29.5 & 65.1 & 28.6 & 51.9 & {54.6} & 82.8 & 7.8 & 57.4\\
DAN ~\cite{long2015learning} & {87.1} & 63.0 & 76.5 & 42.0 &  90.3 & {42.9} & {\bf 85.9} & 53.1 & 49.7 & 36.3 & {\bf 85.8} & 20.7 & 61.1 \\
SimNet ~\cite{pinheiro2018unsupervised} &  {\bf 94.5} & {80.2} & 69.5 & 43.5 & 89.5 & 16.6 & 76.0 & {81.1} & 86.4 & {\bf 76.4} & 79.6 & {41.9} & 69.6\\
MCD ~\cite{saito2018maximum} & 87.0 & 60.9 & {\bf 83.7} & 64.0 & 88.9 & {\bf 79.6} & {84.7} & {76.9} & {88.6} & 40.3 & {83.0} & {25.8} & {71.9}\\
\hline 

LPJT [Ours] & {93.0} & {\bf 80.3} & {66.5} & {56.3} & {\bf 95.8} & {70.3} & {74.2} & {\bf 83.8} & {\bf 91.7} & {40.0} &{78.7} & {\bf 57.6} & {\bf 74.0}\\
\Xhline{0.75pt}
\end{tabular}
\vspace{-10pt}
\end{table*}

\subsection{Model Analysis}
In this section, we analyze our approach by discussing the parameter sensitivity, model convergence, the MMD distance and the landmark selection.

Fig.~\ref{fig:landmarks} illustrates the visual results of landmark selection. By comparing the samples with bigger weights and samples with smaller weights, we can easily capture that the selected landmarks are more similar to the target samples, which proves that our method is able to leverage the most relevant samples for adaptation and filter out outliers to avoid negative transfer.

The parameters sensitivity are reported in Fig.~\ref{fig:paras}. Specifically, Fig.~\ref{fig:paras}(a) shows the effect of locality preservation terms. It can be seen that the value of $\gamma$ should be set as a small value, e.g., from $10^{-5}$ to $10^{-3}$, on Office+Caltech dataset. Fig.~\ref{fig:paras}(a) reports the impact of target variance term. The best values depend on different tests, but it is generally safe to be selected from [10$^{-3}$,1]. Fig.~\ref{fig:paras}(c) shows the influence of the latent space dimensionality. It is obvious that our model is not sensitive to the parameter $d$ on the tested evaluations.

Fig.~\ref{fig:convergance}(a) reports the convergence curve of our approach. We can see that the value of our objective function is monotonically decreasing with the increase of iterations. The convergence curve becomes steady after about $5$ iterations, which proves that our objective converges fast. Fig.~\ref{fig:convergance}(b) shows the MMD distance between the source domain and the target domain. It can be observed that the distribution divergence is getting smaller and smaller with the optimization iterations. In other words, our approach can effectively reduce the discrepancy between the two domains.

\section{Conclusion}
In this paper, we proposed a novel domain adaptation approach for transfer learning. Our approach jointly aligned the source domain and the target domain at both feature level and sample level. At the feature level, we matched the domain distributions by minimizing marginal and conditional MMD. At the sample level, we re-weighted instances by landmarks selection, so that the pivot samples can be selected as the bridge of knowledge transfer and the outliers can be filtered out. At the same time, we considered the locality preservation during knowledge transfer. We also maximized the utilization of sample local relationships by deploying label propagation. Extensive experiments on five visual datasets verified the superiority of our proposed method.

In the experiments, we not only proved that our approach is applicable for both homogeneous and heterogeneous domain adaptation in either unsupervised or semi-supervised manner, but also explored that if handcrafted features can be used to reinforce deep features and if different deep features can be used to reinforce each other. The experiment results showed that one deep feature can be further boosted by handcrafted features and other deep features. Considering that inventing a novel backbone network is quite challenging, using transfer learning to boost the existing features proposes a practical and cheap solution for many real-world applications.


\section*{Acknowledgment}
This work was supported in part by the National Natural Science Foundation of China under Grant 61806039 and 61832001, in part by the National Postdoctoral Program for Innovative Talents under Grant BX201700045, in part by the China Postdoctoral Science Foundation under Grant 2017M623006 and in part by Sichuan Department of Science and Technology under Grant 2019YFG0141.


%



\ifCLASSOPTIONcaptionsoff
  \newpage
\fi



%
\bibliographystyle{IEEEtran}
\bibliography{IEEEabrv,mybib}
%

\vspace{-40pt}

\begin{IEEEbiography}[{\includegraphics[width=1in,height=1.25in,clip,keepaspectratio]{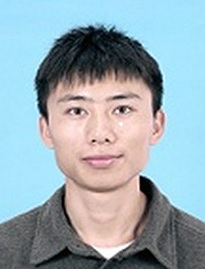}}]{Jingjing Li}
received his MSc and PhD degree in Computer Science from University of Electronic Science and Technology of China in 2013 and 2017, respectively. Now he is a national Postdoctoral Program for Innovative Talents research fellow with the School of Computer Science and Engineering, University of Electronic Science and Technology of China. He has great interest in machine learning, especially transfer learning, subspace learning and recommender systems. 
\end{IEEEbiography}

\vspace{-60pt}
\begin{IEEEbiography}[{\includegraphics[width=1in,height=1.25in,clip,keepaspectratio]{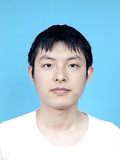}}]{Mengmeng Jing}
received his B.Sc. degree in 2015, and his MSc degree in Computer Application Technology in 2018, both from University of Electronic Science and Technology. He is currently a first year PhD student in the University of Electronic Science and Technology of China. His research interests include machine learning and computer vision.
\end{IEEEbiography}
\vspace{-60pt}

\begin{IEEEbiography}[{\includegraphics[width=1in,height=1.25in,clip,keepaspectratio]{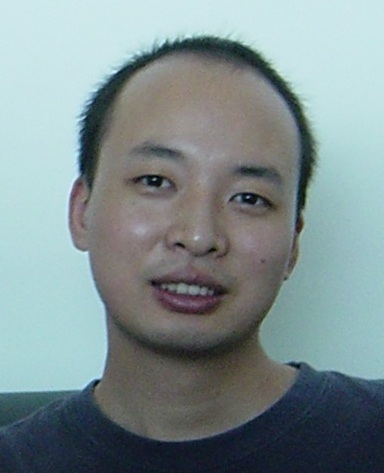}}]{Ke Lu}
received his B.Sc. degree in thermal power engineering from Chongqing University, China in 1996. He obtained his MSc and PhD degrees in Computer Application Technology from the University of Electronic Science and Technology of China, in 2003 and 2006, respectively. He is currently a professor in School of Computer Science and Engineering, University of Electronic Science and Technology of China. His research interests include pattern recognition, multimedia~and~computer~vision.
\end{IEEEbiography}
\vspace{-60pt}

\begin{IEEEbiography}[{\includegraphics[width=1in,height=1.25in,clip,keepaspectratio]{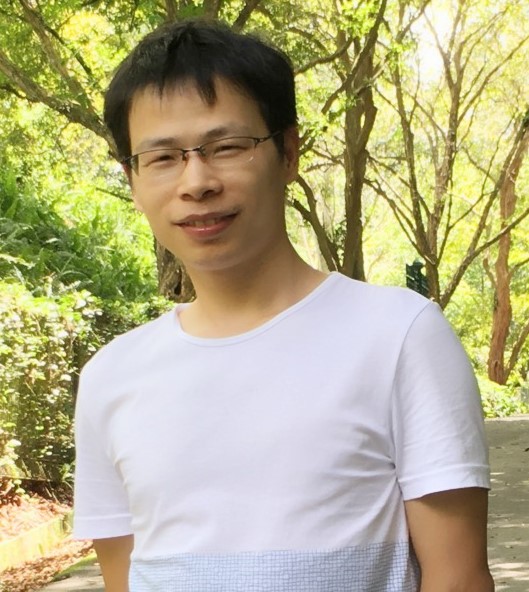}}]{Lei Zhu}
received the B.S. degree (2009) at Wuhan University of Technology, the Ph.D. degree (2015) at Huazhong University of Science and Technology. He is currently a full Professor with the School of Information Science and Engineering, Shandong Normal University, China.  He was a Research Fellow under the supervision of Prof. Heng Tao Shen at the University of Queensland (2016-2017), and Dr. Jialie Shen at the Singapore Management University (2015-2016). His research interests are in the area of large-scale multimedia content analysis and retrieval.
\end{IEEEbiography}
\vspace{-60pt}

\begin{IEEEbiography}[{\includegraphics[width=1in,height=1.25in,clip,keepaspectratio]{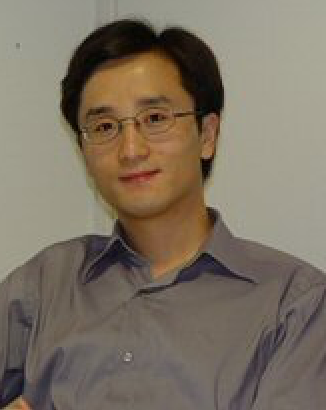}}]{Heng Tao Shen} is currently a Professor of National "Thousand Talents Plan", the Dean of School of Computer Science and Engineering, and the Director of Center for Future Media at the University of Electronic Science and Technology of China. He is also an Honorary Professor at the University of Queensland. He obtained his BSc with 1st class Honours and PhD from Department of Computer Science, National University of Singapore in 2000 and 2004 respectively. He then joined the University of Queensland as a Lecturer, Senior Lecturer, Reader, and became a Professor in late 2011. His research interests mainly include Multimedia Search, Computer Vision, Artificial Intelligence, and Big Data Management. 
\end{IEEEbiography}

\section*{Appendix}
\subsection{Kernelization Extension}
  The proposed method deploys linear projections $A$ and $B$ to learn the subspace. In this section, we show that our method can be easily extended to nonlinear problems by mapping the features into a Reproducing Kernel Hilbert Space (RKHS). Let $\phi(\cdot)$ be a kernel function, $\Phi(X)=[\phi(x_1),\phi(x_2),\cdots,\phi(x_n)]$, the overall objection function Eq.~\eqref{eq:obj_func} can be extended as the following form:
  \begin{equation}  \underset{P,Q}{\mbox{min}}\;{\frac 
        {{\rm Tr}\left(\begin{bmatrix}P^{\rm T}~~ Q^{\rm T}\end{bmatrix}{\begin{bmatrix}M_{\rm ss}+\gamma S^{\rm s}_{\rm w} & M_{\rm su}\\ M_{\rm us} & M_{\rm uu}+\gamma S^{\rm u}_{\rm w}+ \mu I \end{bmatrix}}{\begin{bmatrix}P\\Q\end{bmatrix}}\right)}
        {{\rm Tr}\left({\begin{bmatrix}P^{\rm T} ~~ Q^{\rm T}\end{bmatrix}}{\begin{bmatrix}\gamma S^{\rm s}_{\rm b} & {\bf 0}\\{\bf 0} &\gamma S^{\rm u}_{\rm b}+\mu S^{\rm u}_{\rm h}\end{bmatrix}}{\begin{bmatrix}P\\Q\end{bmatrix}}\right)}} \enspace ,   \label{eq:obj_kernel1} \end{equation} 
  where $P=\Phi(X)A$ and $Q=\Phi(X)B$. For other variables which involves $X_s$ and $X_u$, $X_s$ is replaced by $\Phi(X_s)$ and $X_u$ is replaced by $\Phi(X_u)$. For instance, $M_{ss}$ would be $M_{ss} = \Phi(X_s) ( H_{sm} + H_{sc} ) \Phi(X_s)^T$ in the kernelized version.

  Furthermore, we replace $P$ and $Q$ by $\Phi(X)A$ and $\Phi(X)B$, respectively. The above equation can be rewritten as:
\begin{equation}  \underset{A,B}{\mbox{min}}\;{\frac 
        {{\rm Tr}\left(\begin{bmatrix}A^{\rm T}~~ B^{\rm T}\end{bmatrix}{\begin{bmatrix}M_{\rm ss}+\gamma S^{\rm s}_{\rm w} & M_{\rm su}\\ M_{\rm us} & M_{\rm uu}+\gamma S^{\rm u}_{\rm w}+ \mu K \end{bmatrix}}{\begin{bmatrix}A\\B\end{bmatrix}}\right)}
        {{\rm Tr}\left({\begin{bmatrix}A^{\rm T} ~~ B^{\rm T}\end{bmatrix}}{\begin{bmatrix}\gamma S^{\rm s}_{\rm b} & {\bf 0}\\{\bf 0} &\gamma S^{\rm u}_{\rm b}+\mu S^{\rm u}_{\rm h}\end{bmatrix}}{\begin{bmatrix}A\\B\end{bmatrix}}\right)}} \enspace ,   \label{eq:obj_kernel2} \end{equation}  
  where $S_b^u=K_uL_b^uK_u^T$, $S_b^s=K_sL_b^sK_s^T$, $S_w^u=K_uL_w^uK_u^T$ and $S_w^s=K_sL_w^sK_s^T$ with $K=\Phi(X)^T\Phi(X)$, $K_s=\Phi(X_s)^T\Phi(X_s)$ and $K_u=\Phi(X_u)^T\Phi(X_u)$. Specifically, $S_h^u=\bar{K_u}\bar{K_u}^T$, where $\bar{K_u}={K_u-\mathbf{1}_uK_u-K_u\mathbf{1}_n+\mathbf{1}_uK_u\mathbf{1}_n}$. For the MMD matrices, $M_{ss}=K_s(H_{sm}+\sum_{c=1}^{C}H_{hc}^c)$, $M_{uu}=K_s(H_{um}+\sum_{c=1}^{C}H_{uc}^c)$, $M_{su}=K_s(H_{sum}+\sum_{c=1}^{C}H_{suc}^c)$ and $M_{us}=K_s(H_{sum}+\sum_{c=1}^{C}H_{suc}^c)$. 

It can be seen that the kernelized version Eq.~\eqref{eq:obj_kernel2} is very similar with the liner version Eq.~\eqref{eq:obj_func}. Both of them can be optimized by the same method.

\subsection{Multi-domain Extension}
Gong et al.~\cite{gong2013reshaping} and Hoffman
et al.~\cite{hoffman2012discovering} point out that datasets always consist
of multiple domains, and leveraging these multi-domain information can boost 
the performance of domain adaptation models. Therefore, we further present how to
extend our model to handle multi-source domain adaptation problems.
At beginning, suppose the source dataset
consists of two domains denoted by subscript~$1$ and~$2$, respectively, we need to jointly 
minimize the $\mathrm{MMD}$ metric as follows:
\vspace{-15pt}
\begin{equation}
\label{eq:mmdmm}
  \begin{array}{c}
  \! \\ \bigg\|\frac{1}{n_{s1}}\sum\limits_{i = 1}^{n_{s1}} A^T_1
  x_{s1}^i - \frac{1}{n_u}\sum\limits_{j = 1}^{n_u}B^T_t {x}_{u}^j\bigg\|^2_F +
   \! \\ \bigg\|\frac{1}{n_{s2}}\sum\limits_{i = 1}^{n_{s2}} A^T_2
    {x}_{s2}^i - \frac{1}{n_u}\sum\limits_{j = 1}^{n_u}B^T {x}_{u}^j\bigg\|^2_F ~~,
  \end{array} 
  \vspace{-3pt}
\end{equation}
then, it can be rewritten as:
\begin{equation}
\label{eq:formu2}
  \begin{array}{c}
  \mathrm{Tr}(P^\top X M X^\top P),
  \end{array} 
\end{equation}
where $X$ and $P$ are defined as:
\begin{small}
$$
  X=
  \left(
   \begin{array}{cccc}
            X_{s1} & {0} & {0} & {0}\\
            {0}   & X_{s2} & {0} & {0}       \\
            {0}   & {0} & X_u & {0}\\
            {0}   & {0} & {0} & X_u 
   \end{array}
  \right),
  ~P=[A_1;A_2;B;B],
  $$
  \end{small}
and the MMD matrix ${M}$ can be computed as follows:
\begin{equation}
\begin{array}{l}
 $${M}_{ij}=
\begin{cases}
\frac{1}{n_{s1} n_{s1}}& \textit{, if ${x}_i, {x}_j \in {X}_{s1}$} \\
\frac{1}{n_{s2} n_{s2}}& \textit{, if ${x}_i, {x}_j \in {X}_{s2}$} \\
\frac{1}{n_u n_u}& \textit{, if ${x}_i, {x}_j \in {X}_u$} \\
\frac{-1}{n_{s1} n_u} & \textit{, if ${x}_i \in {X}_{s1/u}, {x}_j \in {X}_{u/s1}$}\\
\frac{-1}{n_{s2} n_u} & \textit{, if ${x}_i \in {X}_{s2/u}, {x}_j \in {X}_{u/s2}$} \\
0& \textit{, otherwise}\\
\end{cases}\\$$
\end{array}
\label{eq:m1}
\vspace{-10pt}
\end{equation}

With the same manner, we can extend the other parts in the formulation into multi-source domain adaptation scenarios. However, the equations would be more complex with the increase of domains. An alternative solution for multi-source domain adaptation has been reported in our previous work CLGA~\cite{liu2018coupled} which addresses multi-source domain adaptation from both global level and local level. At the global level, CLGA regards multiple domains as a unity, and jointly mitigates the gaps of both marginal and conditional distributions between source and target dataset. At the local level, CLGA investigates the relationship among distinctive domains, and exploits both class and domain manifold structures embedded in data samples.

The other parts in our objective function can also be rewritten as the same form 
introduced in Section~\ref{sec:formulation}. Obviously, we can extend
our model to handle more domains, either in the source or the target dataset, in
the same manner.
\end{document}